\def\paperID{0000}
\def\confName{CVPR}
\def\confYear{2025}

\newif\ifreview 
\newif\ifarxiv \newcommand{\arxiv}{\arxivtrue}
\newif\ifcamera 
\newif\ifrebuttal 

\arxiv

\pdfoutput=1
\documentclass[10pt,twocolumn,letterpaper]{article}
\ifreview \usepackage[review]{cvpr} \fi
\ifarxiv \usepackage[pagenumbers]{cvpr} \fi
\ifrebuttal \usepackage[rebuttal]{cvpr} \fi
\ifcamera \usepackage{cvpr} \fi

\usepackage{graphicx}	
\usepackage{amsmath}	
\usepackage{amssymb}	
\usepackage{booktabs}
\usepackage{times}
\usepackage{microtype}
\usepackage{epsfig}
\usepackage{caption}
\usepackage{float}
\usepackage{placeins}
\usepackage{color, colortbl}
\usepackage{stfloats}
\usepackage{enumitem}
\usepackage{tabularx}
\usepackage{xstring}
\usepackage{multirow}
\usepackage{xspace}
\usepackage{url}
\usepackage{subcaption}
\usepackage{xcolor}
\usepackage[hang,flushmargin]{footmisc}

\ifcamera \usepackage[accsupp]{axessibility} \fi

% ...

\ifarxiv  \fi

\newcommand{\R}[1]{{%
    \textbf{%
        \ifstrequal{#1}{1}{\textcolor{red}{R#1}}{%
        \ifstrequal{#1}{2}{\textcolor{blue}{R#1}}{%
        \ifstrequal{#1}{3}{\textcolor{magenta}{R#1}}{%
        \ifstrequal{#1}{4}{\textcolor{teal}{R#1}}{%
                           \textcolor{cyan}{R#1}%
        }}}}%
    }%
}}

\usepackage{xr-hyper}

\makeatletter
\newcommand*{\addFileDependency}[1]{
  \typeout{(#1)}
  \@addtofilelist{#1}
  \IfFileExists{#1}{}{\typeout{No file #1.}}
}

\makeatother
\newcommand*{\myexternaldocument}[1]{
    \externaldocument{#1}
    \addFileDependency{#1.tex}
    \addFileDependency{#1.aux}
}

\definecolor{cvprblue}{rgb}{0.21,0.49,0.74}
\usepackage[pagebackref,breaklinks,colorlinks,allcolors=cvprblue]{hyperref}
\usepackage{bbding}
\usepackage[capitalize]{cleveref}
\crefname{section}{Sec.}{Secs.}
\crefname{table}{Table}{Tables}
\crefname{figure}{Fig.}{Figs.}

\ifarxiv \crefname{appendix}{App.}{Apps.}
\else \crefname{appendix}{Suppl.}{Suppls.} \fi

\def\paperID{11035} 
\def\confName{CVPR}
\def\confYear{2025}

\unless\ifarxiv \myexternaldocument{_supplementary} \fi

\begin{document}
\title{Infinity-MM: Scaling Multimodal Performance with Large-Scale  and High-Quality Instruction Data}

\author{Shuhao Gu\textsuperscript{\rm 1}\thanks{ Core contributors with equal contributions.}, Jialing Zhang\textsuperscript{\rm 1,2*}, Siyuan Zhou\textsuperscript{\rm 1,3*}, Kevin Yu\textsuperscript{\rm 1,4*}, Zhaohu Xing\textsuperscript{\rm 1,5}, Liangdong Wang\textsuperscript{\rm 1}, \\ Zhou Cao\textsuperscript{\rm 1},  Jintao Jia\textsuperscript{\rm 1,4}, Zhuoyi Zhang\textsuperscript{\rm 1,4}, Yixuan Wang\textsuperscript{\rm 1,4}, Zhenchong Hu\textsuperscript{\rm 1,4}, Bo-Wen Zhang\textsuperscript{\rm 1},  \\  Jijie Li\textsuperscript{\rm 1},   Dong Liang\textsuperscript{\rm 1}, Yingli Zhao\textsuperscript{\rm 1}, Songjing Wang\textsuperscript{\rm 1}, Yulong Ao\textsuperscript{\rm 1}, Yiming Ju\textsuperscript{\rm 1}, Huanhuan Ma\textsuperscript{\rm 1}, \\ Xiaotong Li\textsuperscript{\rm 1,6}, Haiwen Diao\textsuperscript{\rm 1,7},  Yufeng Cui\textsuperscript{\rm 1}, Xinlong Wang\textsuperscript{\rm 1}, Yaoqi Liu\textsuperscript{\rm 4}, Fangxiang Feng\textsuperscript{\rm 3}, Guang Liu\textsuperscript{\rm 1}\thanks{Project Lead, \texttt{liuguang@baai.ac.cn}} \\ \\
\textsuperscript{\rm 1}BAAI, \textsuperscript{\rm 2}BJTU, \textsuperscript{\rm 3}BUPT,
\textsuperscript{\rm 4}ICT/CAS,
\textsuperscript{\rm 5}HKUST(GZ),
\textsuperscript{\rm 6}PKU,
\textsuperscript{\rm 7}DLUT}

\maketitle

\begin{abstract}

Recently, Vision-Language Models (VLMs) have achieved remarkable progress in multimodal tasks, and multimodal instruction data serves as the foundation for enhancing VLM capabilities.
Despite the availability of several open-source multimodal datasets, limitations in the scale and quality of open-source instruction data hinder the performance of VLMs trained on these datasets, leading to a significant gap compared to models trained on closed-source data.
To address this challenge, we introduce \textbf{Infinity-MM}, a large-scale multimodal instruction dataset. 
We collected the available multimodal instruction datasets and performed unified preprocessing, resulting in a dataset with over 40 million samples that ensures diversity and accuracy.
Furthermore, to enable large-scale expansion of instruction data and support the continuous acquisition of high-quality data, we propose a synthetic instruction generation method based on a tagging system and open-source VLMs. 
By establishing correspondences between different types of images and associated instruction types, this method can provide essential guidance during data synthesis.
Leveraging this high-quality data, we have trained a 2-billion-parameter Vision-Language Model, \textbf{Aquila-VL-2B}, which achieves state-of-the-art (SOTA) performance among models of similar scale. The data is available at: https://huggingface.co/datasets/BAAI/Infinity-MM. 
\end{abstract}

\section{Introduction}

Recently, Vision-Language Models (VLMs)~\cite{0008LSH23,liu2024improved,Dai0LTZW0FH23,Zhu0SLE24,abs-2308-12966,abs-2311-03079,0004WXDHL00Y24,gpt4v_system_card,yao2024minicpmvgpt4vlevelmllm,DBLP:journals/corr/abs-2409-12191,chen2024far,DBLP:journals/corr/abs-2408-03326} have made significant progresses and found important applications across numerous fields, drawing increasing attention. 
With the development in foundational language models, multimodal architectures, multimodal training data, and evaluation benchmarks, the capabilities of multimodal models have greatly improved. 
Among these elements, multimodal instruction data forms the foundation of a multimodal model’s capabilities.
Expanding training data, enhancing data quality, and increasing data diversity play crucial roles in advancing model performance~\cite{liu2024improved,LiuLWL23a,DBLP:journals/corr/abs-2406-16860,DBLP:journals/corr/abs-2408-03326,li2024llava}.

\begin{figure}[t]
    \centering
    \includegraphics[width=0.95\columnwidth]{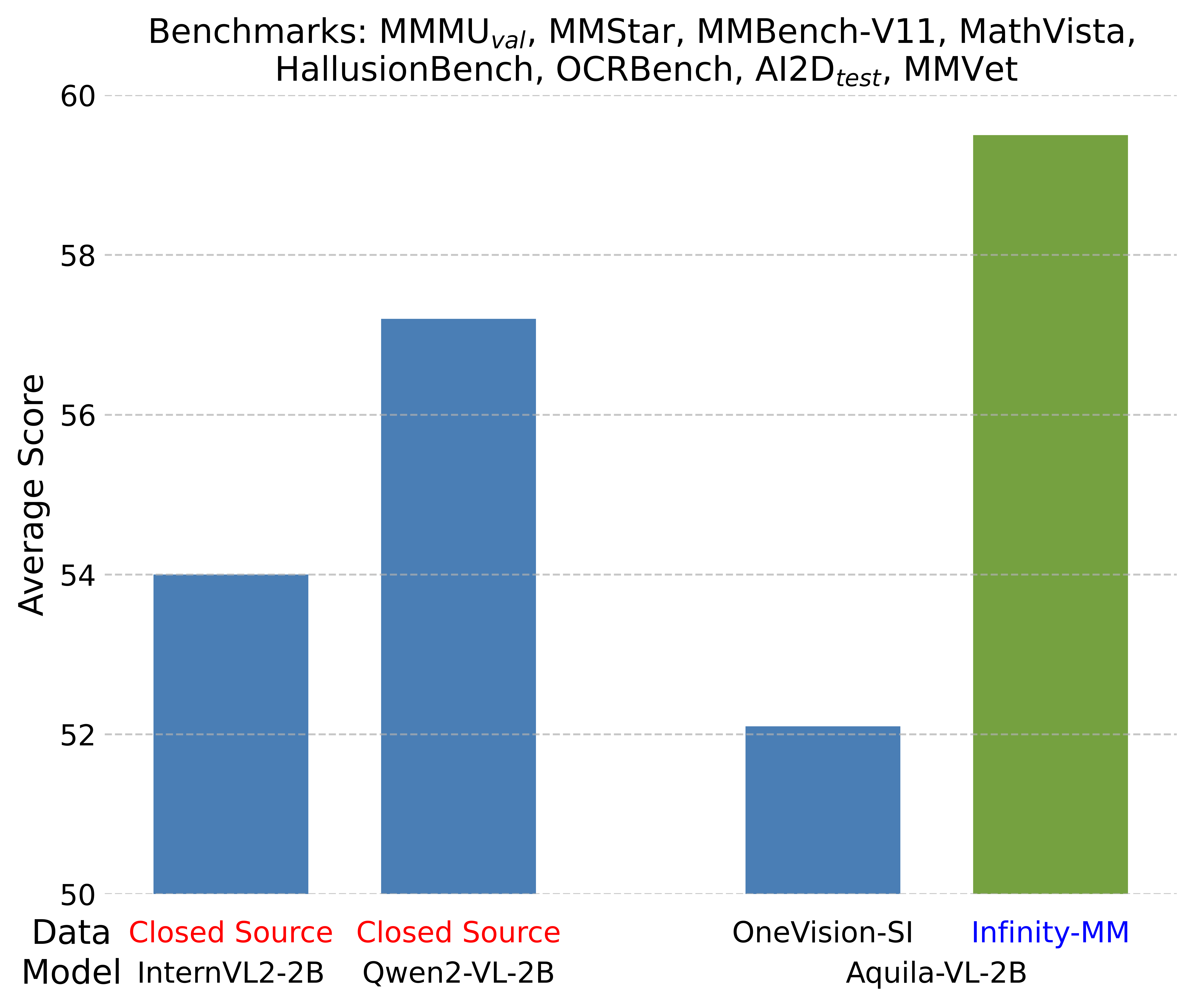}
    \caption{Average score of different VLMs on benchmarks. The Aquila-VL-2B model, trained with Infinity-MM, not only outperforms models trained on other open-source datasets (OneVision-SI) but also surpasses models trained on closed-source datasets.}
    \label{fig:performance}
\end{figure}

Many works have focused on exploring more effective ways to generate and utilize training data. 
For instance, \citet{LiuLWL23a} leverages GPT-4 to generate various types of instructions based on textual descriptions of images.
Building on this, \citet{li2024llava} further expands the data scale, leading to performance improvements. 
\citet{DBLP:journals/corr/abs-2406-16860} and \citet{DBLP:journals/corr/abs-2408-03326} enhance model performance by increasing the dataset size and adjusting the data type ratios.
Besides, several works explore using GPT-4 series models to generate synthetic instruction data, such as captions~\cite{chen2023sharegpt4v,chen2024allava,abs-2407-08303} , OCR data~\cite{textocr-gpt4v}, math questions~\cite{DBLP:journals/corr/abs-2406-17294}, and conversation data~\cite{abs-2311-07574,chen2024allava,DBLP:conf/naacl/LiuWYCSCYY24,DBLP:conf/cvpr/YeXYYHL0Z024}. 
Despite these advancements, existing open-source instruction datasets remain insufficient to support models in achieving best performance. 
Models trained solely on open-source data~\cite{DBLP:journals/corr/abs-2406-16860,DBLP:journals/corr/abs-2408-03326} still significantly lag behind SOTA closed-source models~\cite{gpt4v_system_card,claude} or open-source models trained on closed-source  data~\cite{DBLP:journals/corr/abs-2409-12191,chen2024far,yao2024minicpmvgpt4vlevelmllm}.
Compared to closed-source data, existing open-source data is significantly limited in scale, restricting the foundational capability enhancement of models~\cite{DBLP:journals/corr/abs-2409-05840}. 
Besides, current methods for acquiring high-quality open-source data often rely on commercial closed-source models, such as GPT-4v and GPT-4o, which can be costly, making large-scale data generation a challenge. 
Furthermore, the synthesized instruction types are typically confined to predefined categories or are generated freely by the model without sufficient guidance on instruction types, leading to limitations in correctness and diversity. 
The limitations in both the quantity and quality of open-source data are key factors constraining open model performance.

\begin{table*}[t]
\caption{The comparision between Infinity-MM and other multimodal instruction datasets. "Size" denotes the overall volume of data in the dataset, "Data Synthesis Method" denotes to the specific approach used to generate synthetic data within the dataset, and "Data Composition" denotes the types of instruction data included in the dataset. Compared to existing open-source datasets, Infinity-MM has a significant advantage in data scale. Besides, this is the first time to use an open-source VLM for large-scale, high-quality instruction data synthesis.}
\resizebox{\textwidth}{!}{

\begin{tabular}{c|c|cc|ccccc}
\toprule
\multirow{2}{*}{Datasets}     & \multirow{2}{*}{Size} &  \multicolumn{2}{c|}{Data Synthesis Methods}     & \multicolumn{5}{c}{Data Composition}                                                \\
              &      &      from GPT4(o/v) &
  from Open VLM    & General Instruction & OCR    & Doc/Chart/Screen & Math/Reasoning & Text Instruction \\ \midrule
LVIS-Instruct\citep{DBLP:conf/cvpr/GuptaDG19} & 223K & \CheckmarkBold              & \XSolidBrush           & \CheckmarkBold              & \XSolidBrush & \XSolidBrush           & \XSolidBrush         & \XSolidBrush           \\
Sharegpt4\citep{chen2023sharegpt4v}     & 3.2M & \CheckmarkBold              & \XSolidBrush             & \CheckmarkBold              & \XSolidBrush & \XSolidBrush           & \XSolidBrush         & \XSolidBrush           \\
MMC-INST\citep{DBLP:conf/naacl/LiuWYCSCYY24}      & 500K & \CheckmarkBold              & \XSolidBrush                            & \XSolidBrush              & \XSolidBrush & \CheckmarkBold           & \XSolidBrush         & \XSolidBrush           \\
ALLaVA\citep{chen2024allava}        & 1.7M & \CheckmarkBold              & \XSolidBrush                           & \CheckmarkBold              & \XSolidBrush & \XSolidBrush           & \XSolidBrush         & \XSolidBrush           \\
MathV360K\citep{DBLP:journals/corr/abs-2406-17294}     & 360K & \CheckmarkBold              & \XSolidBrush                          & \XSolidBrush              & \XSolidBrush & \XSolidBrush           & \CheckmarkBold         & \XSolidBrush           \\
DocStruct4M\citep{DBLP:conf/emnlp/WangZ0L20}   & 4M   & \XSolidBrush              & \XSolidBrush              & \XSolidBrush              & \CheckmarkBold & \CheckmarkBold           & \XSolidBrush         & \XSolidBrush           \\
DocDownstream \cite{DBLP:conf/cvpr/YeXYYHL0Z024} & 700K & \CheckmarkBold              & \XSolidBrush & \CheckmarkBold & \CheckmarkBold & \CheckmarkBold & \XSolidBrush & \XSolidBrush \\
Llava-1.5\citep{LiuLWL23a}       & 660K  & \CheckmarkBold              & \XSolidBrush & \CheckmarkBold & \CheckmarkBold & \XSolidBrush & \CheckmarkBold & \CheckmarkBold \\
Cambrain-1\citep{DBLP:journals/corr/abs-2406-16860}    & 10M  & \CheckmarkBold              & \XSolidBrush & \CheckmarkBold              & \CheckmarkBold & \CheckmarkBold           & \CheckmarkBold         & \CheckmarkBold           \\
Llava-OneVision\citep{DBLP:journals/corr/abs-2408-03326} & 7.2M  & \CheckmarkBold              & \XSolidBrush                & \CheckmarkBold & \CheckmarkBold & \CheckmarkBold & \CheckmarkBold & \CheckmarkBold \\
Infinity-MM     & 44.8M & \CheckmarkBold              & \CheckmarkBold & \CheckmarkBold & \CheckmarkBold & \CheckmarkBold & \CheckmarkBold & \CheckmarkBold \\ \bottomrule
\end{tabular}
}
\label{tab:data_com}
\end{table*}

To enhance the performance of open-source models, this work explores improving data effectiveness by expanding the scale of instruction data and increasing the diversity of instruction types. 
We have extensively collected existing open-source multimodal instruction data, constructing a dataset of over 40 million samples, and applied rigorous quality filtering and deduplication processes. 
Besides, we propose a data synthesis method based on a tagging system. 
We establish tagging systems for images and instructions, modeling the relationship between image and instruction types through this system. 
This provides essential guidance for data synthesis, ensuring both the accuracy and diversity of the synthesized data. 
We also employed open-source VLMs for large-scale, high-quality data synthesis, providing detailed and diverse annotations for images to cover all the information within them as comprehensively as possible, thereby enhancing the model's performance.
Ultimately, we successfully trained a 2-billion-parameter model Aquila-VL-2B based on the proposed dataset Infinity-MM, which not only outperforms models trained on other open-source datasets but also surpasses models trained on closed-source datasets, as shown in Figure~\ref{fig:performance}.
The key contributions are summarized as follows:

\begin{itemize}
    \item We collected and organized large-scale multimodal instruction dataset, \textbf{Infinity-MM}, consisting of tens of millions of samples. 
    Through quality filtering and deduplication, we ensured the dataset’s high quality and diversity. 
    \item We propose a data synthesis method based on a tagging system, which provides essential guidance for data synthesis. We employed open-source VLMs to perform large-scale, high-quality instruction data synthesis.
    \item Based on Infinity-MM, we have successfully trained a 2-billion-parameter VLM model, \textbf{Aquila-VL-2B}, achieving SOTA performance among models of the same scale.
\end{itemize}

\section{Related Work}
\noindent \textbf{Multi-modal Instruction Data} Currently, the construction of multimodal instruction datasets primarily employs three methods: manual annotation, model synthesis, and open-source data collection.
The first method is manual annotation.
\citet{DBLP:conf/cvpr/DasKGSYMPB17} involves two annotators per image engaging in up to 10 rounds of question-answer interactions, generating multi-turn dialogues associated with each image.
\citet{DBLP:journals/ci/AgrawalJS24a} employs one set of workers to create up to three QA pairs based on given images and text, followed by verification by a second set of workers to ensure accuracy. 
\citet{DBLP:journals/corr/HavardBR17} combines manual annotation and OCR techniques to build a dataset with detailed text detection and recognition annotations. 
\citet{DBLP:conf/wacv/MathewKJ21} adopts a three-stage manual annotation process to create a visual question-answering dataset tailored for document images, featuring diverse question types and a wide range of document images.
The second kind of methods utilize VLM model to synthesize specific types of data, such as captions~\cite{chen2023sharegpt4v,chen2024allava} , OCR data~\cite{textocr-gpt4v}, math questions~\cite{DBLP:journals/corr/abs-2406-17294}, chart description~\cite{DBLP:conf/cvpr/YeXYYHL0Z024,DBLP:journals/corr/abs-2408-12637}, and conversation data~\cite{abs-2311-07574,chen2024allava,LiuLWL23a}.
The third method is open-source data collection. 
\citet{DBLP:conf/naacl/LiuWYCSCYY24} collected 210K chart-title pairs by downloading academic articles from 2010 to 2020. 
\citet{DBLP:conf/cvpr/YeXYYHL0Z024} aggregates multiple publicly available document image datasets into a unified, diverse document dataset. 
\citet{DBLP:journals/corr/abs-2406-16860} and \citet{DBLP:journals/corr/abs-2408-03326} both compile a variety of available open-source instruction datasets to create a collection encompassing multiple data types.
Our proposed Infinity-MM dataset combines model synthesis with open-source data collection. 
As shown in Table~\ref{tab:data_com}, compared to existing datasets, Infinity-MM includes various types of data and has a substantial advantage in data scale.


\noindent \textbf{Vision-Language Model} VLMs can be categorized into three types based on their capabilities. The first type focuses on understanding multimodal information, such as videos and images~\cite{RadfordKHRGASAM21,AlayracDLMBHLMM22,liu2024improved,0008LSH23,abs-2406-11832}. These models typically take multimodal data as input and produce natural language output, characterized by their ability to integrate and process information from different modalities in a unified manner. 
The second type emphasizes visual generation, primarily aimed at producing high-resolution images and videos~\cite{shi-etal-2020-improving,PeeblesX23,RameshPGGVRCS21,DingYHZZYLZSYT21}. The third type combines both visual understanding and generation capabilities~\cite{SunYCZZWGL0W24,SunCZZYWRL0W24,wang2024emu3nexttokenpredictionneed,abs-2408-11039,abs-2408-12528}. In this work, we focus on enhancing the model's ability to comprehend multimodal information.


\section{The proposed Infinity-MM Dataset}
\begin{table*}[]
\centering
\caption{The data quantity and types contained in Infinity-MM. We segmented the dataset based on data type and quality, allowing the model to perform targeted training at different stages. }
\begin{tabular}{cc|c|c}
\toprule
\multicolumn{2}{c|}{\textbf{Data Category}}                                                       & \textbf{Size} & \textbf{Data Composition}                                                                                                                                       \\ \midrule
\multicolumn{2}{c|}{Image Caption Data}                                               & 10M           & General Instruction Data 10M                                                                                                                                                \\ \midrule
\multicolumn{1}{c|}{\multirow{11}{*}{ Visual Instruction Data}} & Comprehensive Data        & 25.8M         & \begin{tabular}[c]{@{}c@{}}General Instruction Data 7.1M\\  OCR Data 2.6M\\ Doc/Chart/Screen Data 5.8M\\ Math/Reasoning Data 1.3M\\ Text Instruction Data 9M\end{tabular}      \\ \cline{2-4} 
\multicolumn{1}{c|}{}                                         & Selective Data & 6M            & \begin{tabular}[c]{@{}c@{}}General Instruction Data 1.3M\\  OCR Data 0.3M\\ Doc/Chart/Screen Data 1.9M\\ Math/Reasoning Data 0.7M\\ Text Instruction Data 1.8M\end{tabular}    \\ \cline{2-4} 
\multicolumn{1}{c|}{}                                         & GPT4 \& Synthetic Data             & 3M            & \begin{tabular}[c]{@{}c@{}}General Instruction Data 1M\\  OCR Data 0.5M\\ Doc/Chart/Screen Data 0.1M\\ Math/Reasoning Data 0.3M\\ Text Instruction Data 0.3M\\ Our Synthetic Data 0.8M\end{tabular} \\ \bottomrule
\end{tabular}
\label{tab:training_data}
\end{table*}

\begin{figure*}
    \centering
    \includegraphics[width=\textwidth]{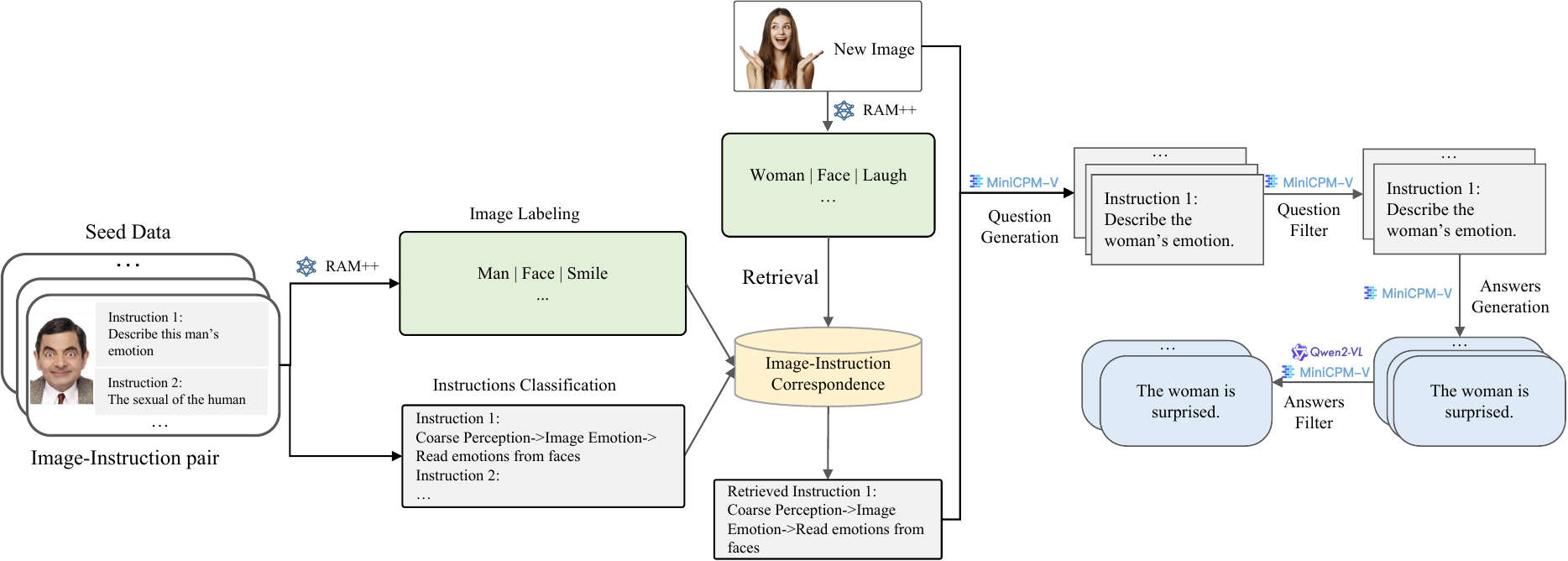}
    \caption{Illustration data synthesis method. We propose a tagging system to model the correspondence between image types and instructions, providing essential guidance for data synthesis to ensure both accuracy and diversity in instruction generation. Besides, we introduce the use of open-source VLM for large-scale, high-quality instruction data synthesis.}
    \label{fig:method}
\end{figure*}

In this section, we provide a detailed explanation of the process to construct the Infinity-MM dataset.
First, we extensively collect available open-source multimodal datasets of various tasks and types and categorize them based on task and quality. 
Then, we propose a data synthesis method based on a tagging system. 
We establish tagging systems for both images and instructions within the multimodal instruction dataset. 
By modeling the relationship between image and instruction tags, we provide essential guidance during instruction synthesis, ensuring that the generated instructions align with the content of the images while also increasing diversity of the synthesized instructions.
Finally, we standardize the formats of data from different sources and apply deduplication and quality filtering to the dataset.

\subsection{Collection of Datasets}

We systematically gather available open-source multimodal datasets and categorize them. These datasets are classified into four categories, as outlined in Table \ref{tab:training_data}. 
The specific sources of the data are given in Appendix~\ref{appen:data}.

\begin{itemize}
[leftmargin=*,itemsep=0pt,topsep=3pt]

\item  \textbf{Image-Caption Data} We collect the Image-Caption dataset generated by Emu2 \citep{SunCZZYWRL0W24}, which performs well in image caption task. The image-caption data has detailed descriptions of image content, making this it well-suited for vision-language alignment training in VLMs.

\item \textbf{Comprehensive Visual Instruction Data} This part of the data is collected from open-source datasets and includes various types of data. The data distribution has not been specifically adjusted, and its effectiveness has not been rigorously validated. Therefore, this part of data is suitable for enhancing the foundational capabilities of VLMs.

\item \textbf{Selective Visual Instruction Data} 
This part of the data is also sourced from open-source datasets, with an adjusted data distribution that increases the proportion of tasks involving mathematical reasoning, chart analysis, and complex instructions—areas where VLMs generally perform poorly.
The quality and effectiveness of these data have been validated, making them suitable for enhancing the instruction-following ability of VLMs.

\item \textbf{GPT4 \& Synthetic Visual Instruction Data} 
We collect instruction data synthesized using GPT-4 series models. 
This data generally exhibits high quality, but the task types are relatively homogeneous, and the data volume is limited. 
We randomly sample a subset of images from this data and synthesized instructions based on these images with the method introduced in Section \ref{synthetic_data_generation}.


\end{itemize}

\subsection{Data Synthesis}

\label{synthetic_data_generation}

In this section, we introduce our multimodal instruction data synthesis method. 
We aim that the generated instructions are closely aligned with the content of the images while maintaining diversity in instruction types and ensuring the accuracy of instruction responses. 
The overall process of the method is shown in Figure~\ref{fig:method}. 
The images of the synthetic data are extracted from the instruction dataset synthesized using the GPT-4 series models, which are of high quality.
However, the instruction type and data quantity of the original data are limited. 
Therefore, we aim to leverage open-source models to synthesize more high-quality data, combining it with the original data to further enhance model performance.

\subsubsection{Image and Instruction Tagging System}
We randomly select a portion of the open-source data as seed data.
Then we utilize the RAM++ model~\cite{huang2023open} to automatically annotate images by extracting key information such as objects, actions, and scenes. These tags form the semantic foundation of the images, providing a critical basis for subsequent instruction generation. The RAM++ model demonstrates excellent performance when processing large-scale image datasets, accurately capturing essential details in multimodal scenes. This lays a solid foundation for generating precise and contextually relevant multimodal instructions.

To systematize the instruction generation process, we design a three-level instruction tagging system that covers different types of instructions. 
Following~\citet{DBLP:journals/corr/abs-2307-06281}, the first-level tags of the instruction tagging system are divided into six categories, which are:
\begin{itemize}
    \item Coarse Perception
    \item Fine-grained Perception (single-instance)
    \item Fine-grained Perception (cross-instance)
    \item Relation Reasoning
    \item Attribute Reasoning
    \item Logic Reasoning
\end{itemize}
We extend the tagging system based on the first-level tags with the GPT-4 model.
The middle level further refines task characteristics, while the bottom level provides a detailed classification based on specific task requirements. 
After constructing the initial tagging system, we use the GPT-4o model to annotate the instructions in the seed data. 
During the annotation process, we refined the tagging system by adding or removing tags.
The resulting tagging system covers 199 sub-tasks, ensuring its comprehensiveness and rationality.
The complete tagging system can be found in Appendix~\ref{appen:label}.



\begin{figure*}
    \centering
    \includegraphics[width=1\textwidth]{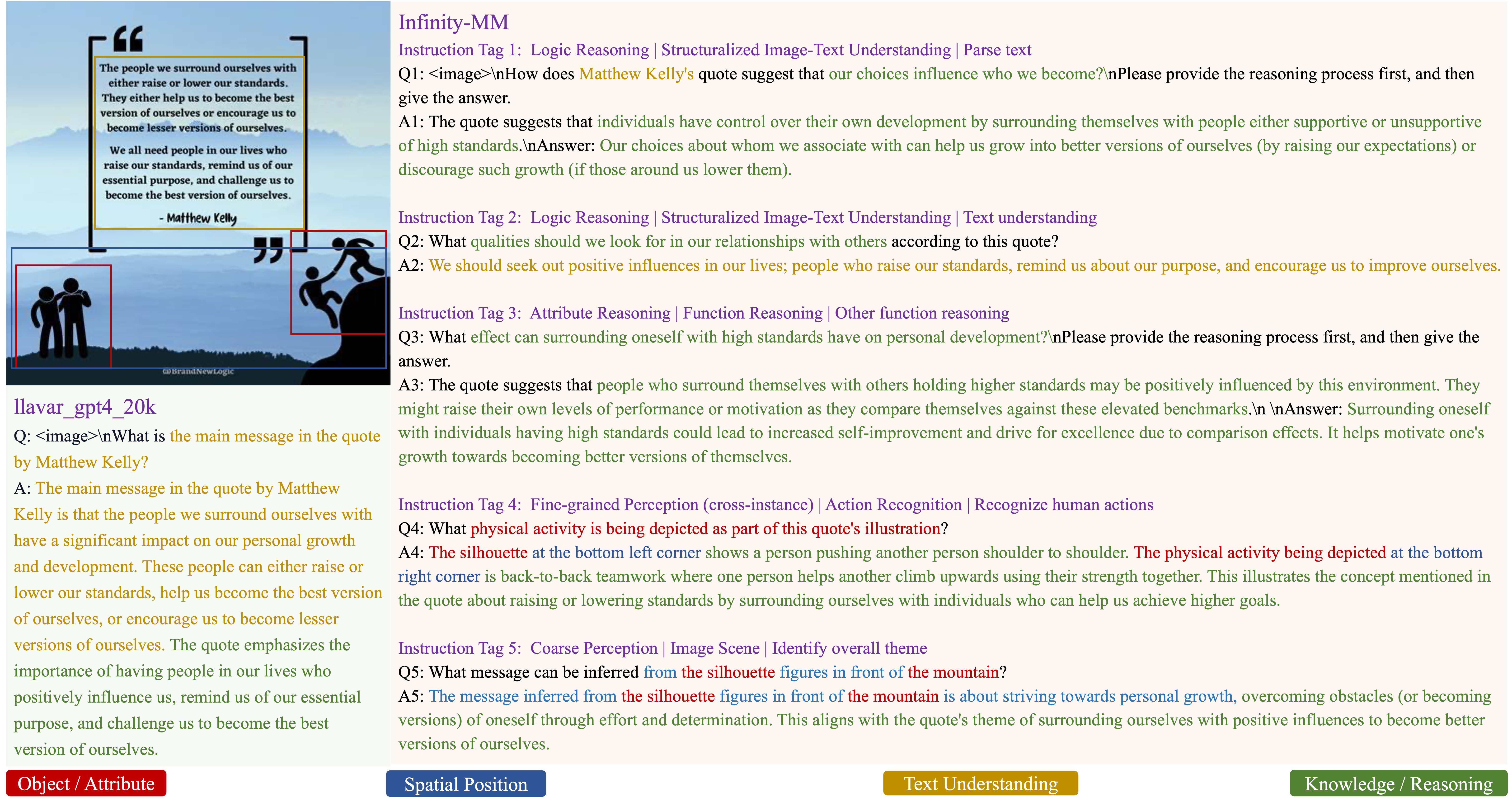}
    \caption{This is an example of our synthesized data. Compared to the original data on the left, our synthesized data on the right offers greater diversity in instructions (Q\&A 1-3) and covers a broader range of image content (Q\&A 4,5).}
    \label{fig:example}
\end{figure*}
\subsubsection{Image-Instruction Mapping}

We annotate both the images and instructions in the seed data with the method in the previous section. 
Following this, we establish the mapping relationships between image tags and instruction tags.
Specifically, we begin by counting the frequency of image tags associated with each instruction tag in the seed data. Then, treating each instruction tag as a distinct unit, we calculate the TF-IDF values for the image tags within these units.
A higher TF-IDF value for an image tag within a given instruction type indicates that images with this tag are more suitable for generating that specific type of instruction. 
Leveraging these TF-IDF-based mappings allows us to automatically determine the most appropriate instruction type to generate for new images. This approach significantly improves the alignment between the generated instructions and the actual image content.

\subsubsection{Question Generation}


After establishing the mapping relationships between images and instruction types, we proceed with instruction synthesis. 
Balancing data synthesis quality and efficiency, we employ the MiniCPM-V2.6 model~\cite{yao2024minicpmvgpt4vlevelmllm} in this work. 
First, we need to generate appropriate questions for each image. 
For each candidate image, we identify the suitable instruction types. 
Then, we input both the image and the target instruction type into the VLM model, prompting it to generate questions based on these information. 
Besides, we select two data examples of the same instruction type from our annotated seed data and input their images and questions into the VLM for reference, enabling few-shot generation. 
For the questions generated by the VLM, we further re-input both the image and question into the VLM to evaluate the relevance of each question to the image, filtering out lower-quality questions.

\subsubsection{Answer Generation}
\begin{figure}
    \centering
    \includegraphics[width=1\linewidth]{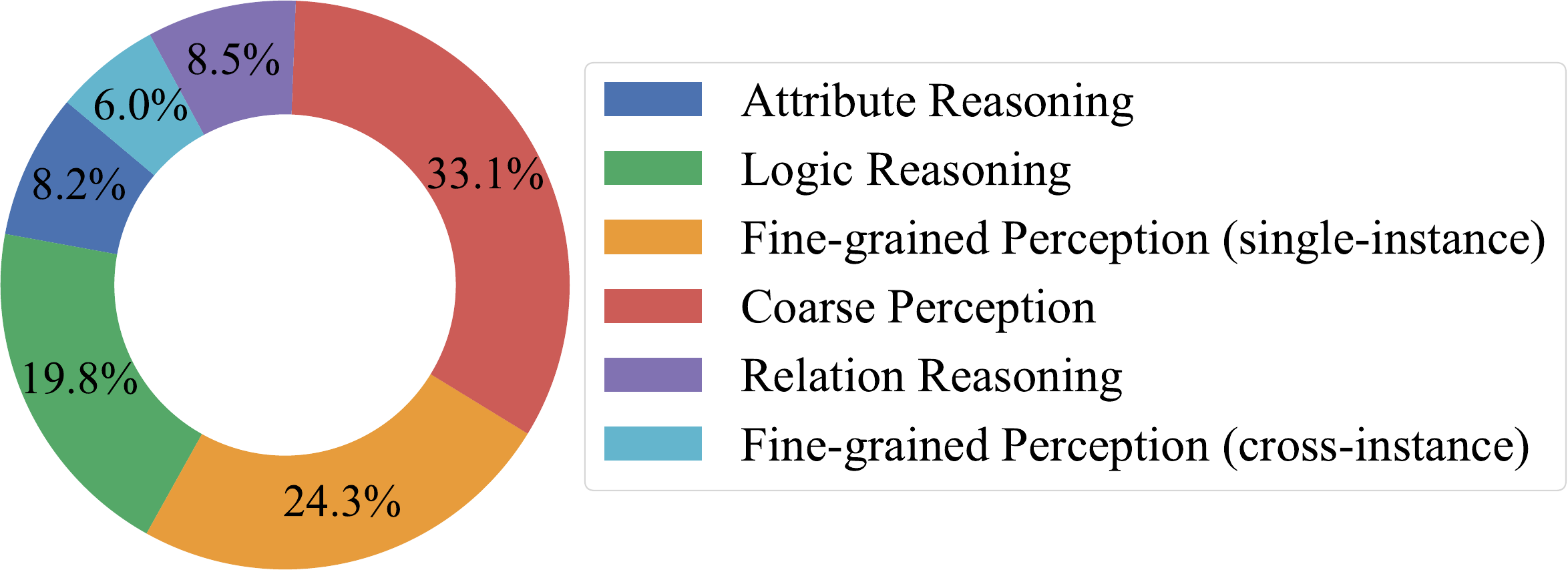}
    \caption{The distribution of first-level tags for synthetic instruction data. The tagging system enables more effective targeted synthesis and analysis of instruction data.}
    \label{fig:type}
\end{figure}

After generating the questions, we proceed to generate the corresponding instruction answers. 
We ensure not only the accuracy of the answers but also the diversity of their formats. 
To achieve this, we employ different prompts during the answer generation process.
Specifically, we employ three different types of prompts: one instructed the model to provide short answers using single words or phrases; another prompted the model to first generate a simple explanation before giving the answer; and the third prompted the model to provide a detailed explanation followed by the answer. 

We re-input the image, question, and answer into the VLM model, allowing it to score the instruction quality on a scale of 1 to 10, where a higher score indicates better quality. 
Instructions with scores below 8 were filtered out. Additionally, we input the synthesized instruction data into the Qwen2-VL-7B model~\cite{DBLP:journals/corr/abs-2409-12191} model to compute the loss, filtering out data with excessively high loss values. 
Finally, we select approximately 3 million filtered data points for training, prioritizing data related to reasoning and document analysis, as these tasks were relatively scarce in the previous training phases.
We combine multiple QA pairs corresponding to the same image into multi-turn instruction data, resulting in approximately 800K training instructions.
Figure~\ref{fig:example} presents an example of our synthesized data.
The distribution of instruction types in the final synthetic data is shown in Figure~\ref{fig:type}.

\begin{table*}[th]
\caption{Configuration for training Aquila-VL-2B across various stages.}
\resizebox{\textwidth}{!}{
\begin{tabular}{@{}clcccccc@{}}
\toprule
\multicolumn{2}{c}{\multirow{2}{*}{}} & \multirow{2}{*}{\textbf{Stage-1}} & \multicolumn{3}{c}{\textbf{Stage-2}} & \multirow{2}{*}{\textbf{Stage-3}} & \multirow{2}{*}{\textbf{Stage-4}} \\ \cmidrule(lr){4-6}
\multicolumn{2}{c}{} &  & a & b & c &  &  \\ \midrule
\multirow{2}{*}{\rotatebox[origin=c]{90}{\textit{Vision}}} & \textbf{Resolution} & 384 & 384\footnotesize{$\times$\{(1$\times$1),...,(2$\times$2)\}} & 384\footnotesize{$\times$\{(1$\times$1),...,(3$\times$3)\}} & 384\footnotesize{$\times$\{(1$\times$1),...,(4$\times$4)\}} & 384\footnotesize{$\times$\{(1$\times$1),...,(6$\times$6)\}} & 384\footnotesize{$\times$\{(1$\times$1),...,(6$\times$6)\}} \\
 & \#tokens & 729 & Max 5$\times$729 & Max 6$\times$729 & Max 7$\times$729 & Max 10$\times$729 & Max 10$\times$729 \\ \midrule
\rotatebox[origin=c]{90}{\textit{Data}} & \textbf{Samples} & 10M & 8.6M & 8.6M & 8.6M & 6M & 3M \\ \midrule
\multirow{2}{*}{\rotatebox[origin=c]{90}{\textit{Model}}} & \textbf{Trainable} & Projector & Full Model & Full Model & Full Model & Full Model & Full Model \\
 & \#para\_counts & 4.13M & 2B & 2B & 2B & 2B & 2B \\ \midrule
\multirow{4}{*}{\rotatebox[origin=c]{90}{\;\;\;\;\;\textit{Training}}} & \textbf{Batch Size} & 512 & 512 & 512 & 512 & 512 & 512 \\
 & \textbf{LR} & 1.00E-03 & 1.00E-05 & 1.00E-05 & 1.00E-05 & 1.00E-05 & 1.00E-05 \\
 & \textbf{Epoch} & 1 & 1 & 1 & 1 & 1 & 1 \\ \bottomrule
\end{tabular}}
\label{training_hypara}
\end{table*}

\subsection{Data Processing}

After collecting all the data, we proceed with data processing. First, we standardize the format of data from various sources. Then, we remove duplicate Image-Text pairs and filter out images with high similarity based on their pHash values~\cite{Zauner2010ImplementationAB}. 
Besides, we use Qwen2-VL-2B~\cite{DBLP:journals/corr/abs-2409-12191} to calculate the loss for each sample and exclude the top 5\% data with the highest loss, as high loss in well-trained multimodal models often indicates noisy data. 

    

\section{Training of Aquila-VL-2B}


\subsection{Model Architecture}
To validate the effectiveness of Infinity-MM, we used it to train Aquila-VL-2B, a vision-language model with about 2 billion parameters.
Aquila-VL-2B builds upon LLaVA-OneVision architecture\citep{DBLP:journals/corr/abs-2408-03326}, comprising a language tower, a vision tower, and a projector. 
\begin{itemize}
[leftmargin=*,itemsep=0pt,topsep=5pt]
    \item \textbf{Language Tower} We choose Qwen-2.5-Instruct \citep{qwen} as the language tower for its outstanding performance among open-source models.
    \item \textbf{Vision Tower} We utilize SigLIP \citep{zhai2023sigmoid}, with approximately 400 million parameters, as the vision tower to extract visual features from input images and videos.
    \item \textbf{Projector} We utilize a two-layer MLP \citep{liu2024improved} with a GELU \citep{hendrycks2023gaussianerrorlinearunits} activation to project visual features into the word embedding space.

\end{itemize}

\begin{table*}[t]
\caption{Comprehensive benchmark Comparisons of Aquila-VL-2B Model and other models. Results for models marked with * are sourced directly from the VLMEvalKit leaderboard, while other results are measured by us using default configurations.
In all experimental results, higher values indicate better performance. The Aquila-VL-2B model, trained with Infinity-MM, achieved SOTA performance on average among models of the same scale.}
\centering
\resizebox{\textwidth}{!}{%
\begin{tabular}{l|c|c|c|c|c|c|c|c|c|c|c|c}
\toprule
 & Model  & \small{MiniMonkey}*\cite{huang2024mini} & \small{PaliGemma}*\cite{beyer2024paligemma} & \small{DeepSeekVL}*\cite{lu2024deepseekvl} & H2OVL*\cite{galib2024h2ovlmississippivisionlanguagemodels} & Phi3-V*\cite{abdin2024phi3technicalreporthighly} & Vintern*\cite{Vintern-3B} & \small{MiniCPM-V2}\cite{yao2024minicpmvgpt4vlevelmllm} & InternVL2\cite{chen2024far} & \small{XinYuanVL}\cite{xinyuan} & Qwen2VL\cite{DBLP:journals/corr/abs-2409-12191} & Aquila-VL \\  \cmidrule(lr){2-13}
 & Pars (B) & 2.2 & 2.9 & 2.0 & 2.1 & 4.2 & 3.7 & 2.8 & 2 & 2 & 2 & 2 \\  \cmidrule(lr){2-13}
 & Open Data & \CheckmarkBold & \XSolidBrush & \XSolidBrush & \XSolidBrush & \XSolidBrush & \CheckmarkBold & \XSolidBrush & \XSolidBrush & \XSolidBrush & \XSolidBrush & \CheckmarkBold \\  \midrule
 & MMB-EN\textsubscript{test}\cite{DBLP:journals/corr/abs-2307-06281} & 72.4 & 71.0 & 66.4 & 72.1 & 73.6 & 70.6 & 69.4 & 73.4 & \textbf{78.9} & 74.9 & 78.8 \\
 & MMB-CN\textsubscript{test}\cite{DBLP:journals/corr/abs-2307-06281} & 70.3 & 63.6 & 62.9 & 62.9 & 62.4 & 69.4 & 65.9 & 70.9 & 76.1 & 73.9 & \textbf{76.4} \\
 & MMB\textsubscript{V1.1}\cite{DBLP:journals/corr/abs-2307-06281} & 68.9 & 65.6 & 63.8 & 64.8 & 65.2 & 66.6 & 65.2 & 69.7 & 75.4 & 72.7 & \textbf{75.2} \\
 & MMT-Bench\textsubscript{all}\cite{DBLP:conf/icml/YingMWLLYZZLLLL24} & - & - & - & - & - & - & 54.5 & 53.3 & 57.2 & 54.8 & \textbf{58.2} \\
 & RealWorldQA\cite{grok2024} & 57.1 & 55.2 & 49.7 & 62.9 & 58.8 & 58.2 & 55.4 & 57.3 & 63.9 & 62.6 & \textbf{63.9} \\
 & HalluB\cite{DBLP:conf/cvpr/GuanLWXLL0CHYM024} & 38.7 & 32.2 & 27.6 & 35.9 & 39.0 & \textbf{43.2} & 36.8 & 38.1 & 36.0 & 41.5 & 43.0 \\
 & S-Bench2\textsubscript{Plus}\cite{DBLP:journals/corr/abs-2404-16790} & - & 49.8 & 43.7 & 59.5 & \textbf{64.2} & 64.1 & 51.8 & 60.0 & 63.0 & 62.4 & 63.0 \\
 & LLaVABench\cite{DBLP:conf/nips/LiuLWL23a} & 61.2 & 36.9 & 51.1 & 65.7 & 63.9 & 62.2 & 66.1 & 64.8 & 42.4 & 52.5 & \textbf{68.4} \\
 & MMStar\cite{DBLP:journals/corr/abs-2403-20330} & 50.4 & 48.3 & 39.9 & 48.9 & 47.7 & 47.5 & 41.6 & 50.2 & 51.9 & 47.8 & \textbf{54.9} \\
 & POPE\cite{DBLP:conf/emnlp/LiDZWZW23} & - & 87.5 & 85.9 & 86.6 & 83.7 & 87.4 & 86.6 & 85.3 & \textbf{89.4} & 88.0 & 83.6 \\
 & MMVet\cite{DBLP:journals/corr/abs-2308-02490} & 38.0 & 33.1 & 29.2 & 41.1 & 44.1 & 37.8 & 44.0 & 41.1 & 42.7 & \textbf{50.7} & 44.3 \\
\multirow{-12}{*}{\shortstack{\small{GeneralVQA}}} & MME\cite{DBLP:journals/corr/abs-2306-13394} & 1881.6 & 1686.1 & 1531.6 & 1767.5 & 1508.0 & 1782.6 & 1788.6 & 1863.0 & 1854.9 & \textbf{1890.0} & 1799.3 \\  \midrule
 & MMMU\textsubscript{val}\cite{DBLP:journals/corr/abs-2311-16502} & 35.0 & 34.9 & 33.8 & 36.3 & 46.1 & 46.7 & 39.6 & 34.9 & 43.6 & 41.7 & \textbf{47.4} \\
 & S-QA\textsubscript{test}\cite{DBLP:conf/nips/LuMX0CZTCK22} & - & 94.3 & 68.4 & 92.1 & 90.0 & 75.0 & 80.4 & 94.1 & 86.6 & 78.1 & \textbf{95.2} \\
 & AI2D\textsubscript{test}\cite{DBLP:conf/eccv/KembhaviSKSHF16} & 74.8 & 68.3 & 51.5 & 70.9 & \textbf{78.4} & 69.1 & 64.8 & 74.4 & 74.2 & 74.6 & 75.0 \\
 & MathVista\cite{DBLP:conf/iclr/LuBX0LH0CG024} & 45.2 & 28.7 & 29.8 & 56.5 & 44.6 & 43.4 & 39.0 & 45.0 & 47.1 & 47.9 & \textbf{59.0} \\
 & MathVerse\cite{DBLP:journals/corr/abs-2403-14624} & - & - & - & - & - & - & 19.8 & 24.7 & 22.2 & 21.0 & \textbf{26.2} \\ 
\multirow{-6}{*}{\shortstack{\small{Knowledge\&Math}}} & MathVision\cite{DBLP:journals/corr/abs-2402-14804} & - & - & - & - & - & - & 15.4 & 12.6 & 16.3 & 17.5 & \textbf{18.4} \\  \midrule
 & DocVQA\textsubscript{test}\cite{DBLP:conf/wacv/MathewKJ21} & - & - & - & - & - & - & 71.0 & 86.9 & 87.6 & \textbf{89.9} & 85.0 \\
 & InfoVQA\textsubscript{test}\cite{DBLP:conf/wacv/MathewKJ21} & - & - & - & - & - & - & 40.0 & 59.5 & 59.1 & \textbf{65.4} & 58.3 \\
 & ChartQA\textsubscript{test}\cite{DBLP:journals/corr/abs-2203-10244} & - & 33.7 & 47.4 & 59.4 & \textbf{81.8} & 68.3 & 59.6 & 71.4 & 57.1 & 73.5 & 76.5 \\
 & TextVQA\textsubscript{val}\cite{DBLP:conf/cvpr/SinghNSJCBPR19} & - & 68.1 & 57.8 & 74.8 & 72.4 & 67.2 & 74.3 & 73.5 & 77.6 & \textbf{79.9} & 76.4 \\
 & OCRVQA\textsubscript{test}\cite{DBLP:conf/icdar/0001SSC19} & - & 57.8 & 58.1 & \textbf{70.3} & 61.9 & 56.8 & 54.4 & 40.2 & 67.6 & 68.7 & 64.0 \\
 & VCR\textsubscript{en easy}\cite{DBLP:journals/corr/abs-2406-06462} & - & - & - & - & - & - & 27.6 & 51.6 & 67.7 & 68.3 & \textbf{ 70.0} \\
\multirow{-6}{*}{Text-rich} & OCRBench \cite{DBLP:journals/corr/abs-2305-07895} & 804 & 614 & 413 & 778 & 637 & 618 & 613 & 784 & 782 & \textbf{810} & 772 \\  \midrule
 
 & Average & - & - & - & - & - & - & 53.9 & 59.1 & 61.1 & 62.3 & \textbf{64.1} \\ \bottomrule
\end{tabular}%
}
\label{tab:main_res}
\end{table*}

\subsection{Training Details} 

We use the official codes of LLaVA-OneVision\footnote{\url{https://github.com/LLaVA-VL/LLaVA-NeXT/tree/main/scripts/train}} and the training setup is given in Table \ref{training_hypara}.
Followling~\citet{li2024llava}, the training is divided into different stages, progressively increasing the difficulty, image resolution, and data quality:

\begin{itemize}
[leftmargin=*,itemsep=0pt,topsep=5pt]
    \item \textbf{Stage 1:} 
    We train the projector using 10M image-caption data to align the visual feature space with the word embedding space. Both the vision tower and language tower are frozen during this phase.
    \item \textbf{Stage 2:} We utilize general visual instruction data for further training to equip the model with fundamental capabilities for solving multimodal tasks. The data is divided into three subsets, and during each stage of training, the maximum visual resolution was progressively increased to enhance the model’s comprehension of visual information.
    \item \textbf{Stage 3:} We employ selective visual instruction data for training and further increase the maximum resolution to improve performance.
    \item \textbf{Stage 4:} We fine-tune the model using training data from GPT-4 and synthetic data.
    Experiments demonstrate that this part of data can further enhance model performance.
    
\end{itemize}




\section{Evaluation}


In this section, we first compared the performance of Aquila-VL-2B with similarly sized models across different benchmarks, showing that Aquila-VL-2B achieves SOTA performance. 
Next, we assess the impact of training with Infinity-MM versus other open-source datasets.
Finally, we analyze the impact of our synthetic data on model performance and examine how performance varies with data scale.

\subsection{Main Results}

We assessed the visual capabilities of Aquila-VL-2B using a range of visual benchmarks provided by the VLMEvalKit~\cite{duan2024vlmevalkit}. Experimental results are shown in Table \ref{tab:main_res}. 
Aquila-VL-2B demonstrates highly competitive performance at the same scale, achieving new state-of-the-art results. Specifically, we evaluated the capabilities of Aquila-VL-2B across three task categories.

\noindent \textbf{General Visual Question Answering} We conducted extensive evaluations across a diverse array of general visual question answering benchmarks:
MMBench-EN\textsubscript{test} (MMB-EN\textsubscript{test}), MMBench-CN\textsubscript{test} (MMB-CN\textsubscript{test}), MMBench-1.1 (MMB\textsubscript{V1.1}) \cite{DBLP:journals/corr/abs-2307-06281}, 
MMT-Bench\textsubscript{all} \cite{DBLP:conf/icml/YingMWLLYZZLLLL24}, RealWorldQA \cite{grok2024}, HallusionBench (HalluB) \cite{DBLP:conf/cvpr/GuanLWXLL0CHYM024}, SEEDBench2\textsubscript{plus} (S-Bench2\textsubscript{plus}) \cite{DBLP:journals/corr/abs-2404-16790}), LLaVABench \cite{DBLP:conf/nips/LiuLWL23a},
MMStar \cite{DBLP:journals/corr/abs-2403-20330},POPE \cite{DBLP:conf/emnlp/LiDZWZW23} ,MMVet \cite{DBLP:journals/corr/abs-2308-02490}, and  MME \cite{DBLP:journals/corr/abs-2306-13394} datasets. Aquila-VL-2B demonstrates strong performance across these benchmarks, achieving or surpassing state-of-the-art results in most cases at the same scale.
Specifically, Aquila-VL-2B achieves the best performance on MMB-CN\textsubscript{test}, MMB\textsubscript{V1.1}, MMT-Bench\textsubscript{all}, RealWorldQA, LLaVABench, and MMSar. It also demonstrates top-tier results on MMB-EN\textsubscript{test}, HalluB, and S-Bench2\textsubscript{Plus}. Additionally, it achieves competitive results on POPE, MMVet, and MME, with room for further improvement. These results strongly indicate Aquila-VL-2B's remarkable capability in handling general visual question-answering tasks and validate the effectiveness of using the Infinity-MM dataset to enhance visual understanding.

\noindent \textbf{Knowledge and Mathematical Reasoning} We conducted experiments on MMMU\textsubscript{val} \cite{DBLP:journals/corr/abs-2311-16502}, 
ScienceQA\textsubscript{test} (S-QA\textsubscript{test}) \cite{DBLP:conf/nips/LuMX0CZTCK22},
AI2D\textsubscript{test} \cite{DBLP:conf/eccv/KembhaviSKSHF16}, MathVista \cite{DBLP:conf/iclr/LuBX0LH0CG024},  MathVerse \cite{DBLP:journals/corr/abs-2403-14624}, and MathVision \cite{DBLP:journals/corr/abs-2402-14804} to evaluate the model's capabilities in knowledge and mathematical reasoning.
Aquila-VL-2B achieves state-of-the-art performance on MMMU\textsubscript{val}, S-QA\textsubscript{val}, MathVista, MathVerse, and MathVision, while securing top-tier results on AI2D\textsubscript{test}. These outcomes underscore Aquila-VL-2B's exceptional capabilities in visual knowledge comprehension and mathematical reasoning. 
The success can be attributed to Infinity-MM, which has collected over 2 million high-quality mathematical reasoning data points, further demonstrating the dataset's quality and effectiveness.

\noindent \textbf{Text Reading}
We assessed Aquila-VL-2B's capabilities in text reading and diagram comprehension using DocVQA\textsubscript{test} \cite{DBLP:conf/wacv/MathewKJ21}, InfoVQA\textsubscript{test} \cite{DBLP:conf/wacv/MathewKJ21}, ChartQA\textsubscript{test} \cite{DBLP:journals/corr/abs-2203-10244}, TextVQA\textsubscript{val} \cite{DBLP:conf/cvpr/SinghNSJCBPR19}, OCRVQA\textsubscript{testcore} \cite{DBLP:conf/icdar/0001SSC19} and VCR\textsubscript{en easy} \cite{DBLP:journals/corr/abs-2406-06462} and  OCRBench \cite{DBLP:journals/corr/abs-2305-07895}. 
The experimental results indicate that Aquila-VL-2B’s text comprehension abilities are generally among the top tier, with potential for further improvement. We will continually supplement high-quality, text-rich data to further refine the Infinity-MM dataset.
\begin{table*}[]
\centering
\caption{Performance comparison of models trained with Infinity-MM versus other open-source datasets. Compared to existing open-source datasets, Infinity-MM demonstrates a clear performance advantage.}
\resizebox{1.0\textwidth}{!}{%
\begin{tabular}{l|c|c|c|c|c|c|c|c|c|c|c}
\toprule
Model & Params (B) & Training Data & Average & MMBenchV1.1\textsubscript{test} & MMStar & MMMU\textsubscript{val} & MathVista\textsubscript{testmini} & HallusionBench & AI2D\textsubscript{test} & OCRBench & MMVet \\ \midrule
LLaVA-OneVision-7B~\cite{DBLP:journals/corr/abs-2408-03326} & 8 & LLaVA-OneVision-SI & 60.1 & 80.9 & 61.9 & 47.9 & 62.3 & 31.6 & 82.4 & 622 & 51.9 \\
Cambrian-34B~\cite{DBLP:journals/corr/abs-2406-16860} & 34 & Cambrian-7M & 58.3 & 77.8 & 54.2 & 50.4 & 50.3 & 41.6 & 79.5 & 591 & 53.2 \\ \midrule
\multirow{2}{*}{Aquila-VL-2B} & \multirow{2}{*}{2} & LLaVA-OneVision-SI & 52.1 & 70.3 & 49.9 & 42.2 & 49.4 & 33.6 & 73.1 & 617 & 36.7 \\
 &  & Infinity-MM & 59.5 & 75.2 & 54.9 & 47.4 & 59.0 & 43.0 & 75.0 & 772 & 44.3 \\ \bottomrule
\end{tabular}%
}
\label{tab:model_comparison}
\end{table*}

\subsection{Infinity-MM vs. Other Open Datasets}

To further validate the performance advantages of Infinity-MM and its importance for research in the open-source community, we compared Infinity-MM with several widely used, high-performance open-source datasets. We selected eight of the most representative datasets for testing, and the experimental results are shown in Table \ref{tab:model_comparison}. 
Specifically, we first compare Aquila-VL-2B, trained on Infinity-MM, with LLaVA-OneVision-7B, trained on LlaVA-OneVision data, and Cambrian-34B, trained on Cambrian-10M. The results indicate that, despite its smaller scale relative to LLaVA-OneVision-7B and Cambrian-34B, Aquila-VL-2B outperforms Cambrian-34B and performs comparably to LLaVA-OneVision-7B. This underscores the critical role of training data in determining model performance.
Furthermore, to better assess the relative impact of Infinity-MM versus LLaVA-OneVision data, we trained Aquila-VL-2B using each dataset independently. The results reveal that training on Infinity-MM yields superior performance, providing strong evidence of the high quality of the Infinity-MM dataset.

\subsection{Effects of Synthetic Data}

\begin{table*}[t]
\centering
\caption{Comparison of the model performance with and without synthetic data. 
The average performance of the model significantly declined without synthetic data, validating the effectiveness of synthetic data.}
\resizebox{\textwidth}{!}{%
\begin{tabular}{l|c|c|c|c|c|c|c|c|c}
\toprule
Models  & Average & MMBenchV1.1\textsubscript{test} & MMStar & MMMU\textsubscript{val} & MathVista\textsubscript{testmini} & HallusionBench & AI2D\textsubscript{test} & OCRBench & MMVet \\ \midrule
Aquila-VL-2B & 59.5 & 75.2 & 54.9 & 47.4 & 59.0 & 43.0 & 75.0 & 772 & 44.3 \\ 
\;\;w/o Synthetic Data & 57.25 & 75.09 & 54.53 & 45.56  & 57 & 35.89 & 75.03 & 766 & 38.3 \\
\bottomrule
\end{tabular}%
}
\label{tab:ablation}
\end{table*}



To assess the impact of synthetic data on model performance, we conducted an ablation study. In this experiment, we removed all synthetic data and trained the model using only the original GPT-generated data. The results, as shown in Table~\ref{tab:ablation}, reveal a significant decline in overall model performance after removing the synthetic data. This demonstrates that the synthetic data played a crucial role in enhancing the model’s performance, further validating the effectiveness of our approach in data augmentation and diversity.

\subsection{Data Scaling}

\begin{figure}
    \centering
    \includegraphics[width=1\linewidth]{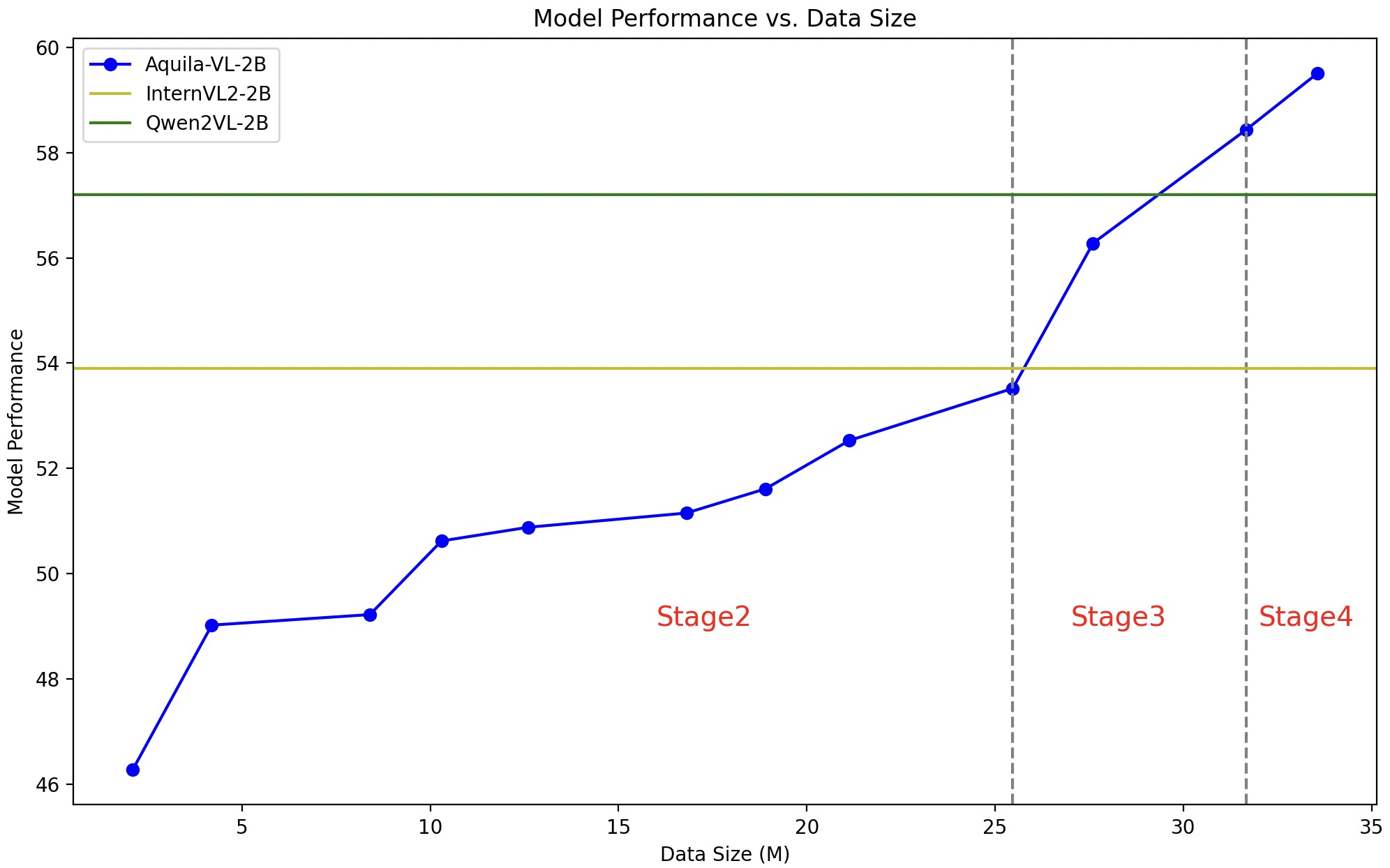}
    \caption{
As the amount of instruction data used for training increased, the model's average performance continued to improve, validating the effectiveness of scaling up instruction data.}
    \label{fig:ds}
\end{figure}

To further analyze the impact of data size scaling on model performance, we conducted a detailed study on how model performance varies with the amount of training data. The results, shown in Figure~\ref{fig:ds}, indicate a consistent improvement in performance as the training data increases. This trend clearly demonstrates that expanding the scale of instruction data has a significant positive effect on model performance.
This observation indicates that incorporating more diverse instructional data enhances the Aquila-VL's capability to tackle complex tasks. Consequently, expanding the volume of instructional data proves to be an effective strategy for boosting overall model performance.

\section{Conclusion}
In this work, we proposed the Infinity-MM multimodal instruction dataset, comprising tens of millions of samples to substantially increase data volume and enhance model effectiveness. 
Furthermore, we introduced a novel method for synthesizing instruction data based on open-source models, which enabled the generation of high-quality instruction data and further expanded the dataset. Finally, we trained the Aquila-VL-2B model on Infinity-MM, achieving state-of-the-art performance for models of a similar scale.

\noindent \textbf{Limitations} 
Despite the impressive performance of Infinitiy-MM, there is still room for enhancement in our work: (1) Due to limited computational resources, we have currently conducted experiments only on the 2B-scale model; (2) Due to time constraints, our dataset has not yet incorporated additional multi-image, video, and multilingual data. 
The future work will involve training on larger-scale models to further validate the quality of Infinity-MM. Additionally, we will continuously incorporate diverse high-quality data into Infinity-MM to support advancements in related research.


{\small
\bibliographystyle{ieeenat_fullname}
\bibliography{custom}

\begin{thebibliography}{80}
\providecommand{\natexlab}[1]{#1}
\providecommand{\url}[1]{\texttt{#1}}
\expandafter\ifx\csname urlstyle\endcsname\relax
  \providecommand{\doi}[1]{doi: #1}\else
  \providecommand{\doi}{doi: \begingroup \urlstyle{rm}\Url}\fi

\bibitem[Abdin et~al.(2024)Abdin, Aneja, Awadalla, Awadallah, Awan, Bach, Bahree, Bakhtiari, Bao, Behl, Benhaim, Bilenko, Bjorck, Bubeck, Cai, Cai, Chaudhary, Chen, Chen, Chen, Chen, Chen, Cheng, Chopra, Dai, Dixon, Eldan, Fragoso, Gao, Gao, Gao, Garg, Giorno, Goswami, Gunasekar, Haider, Hao, Hewett, Hu, Huynh, Iter, Jacobs, Javaheripi, Jin, Karampatziakis, Kauffmann, Khademi, Kim, Kim, Kurilenko, Lee, Lee, Li, Li, Liang, Liden, Lin, Lin, Liu, Liu, Liu, Liu, Liu, Luo, Madan, Mahmoudzadeh, Majercak, Mazzola, Mendes, Mitra, Modi, Nguyen, Norick, Patra, Perez-Becker, Portet, Pryzant, Qin, Radmilac, Ren, de~Rosa, Rosset, Roy, Ruwase, Saarikivi, Saied, Salim, Santacroce, Shah, Shang, Sharma, Shen, Shukla, Song, Tanaka, Tupini, Vaddamanu, Wang, Wang, Wang, Wang, Wang, Wang, Ward, Wen, Witte, Wu, Wu, Wyatt, Xiao, Xu, Xu, Xu, Xue, Yadav, Yang, Yang, Yang, Yang, Yu, Yuan, Zhang, Zhang, Zhang, Zhang, Zhang, Zhang, Zhang, and Zhou]{abdin2024phi3technicalreporthighly}
Marah Abdin, Jyoti Aneja, Hany Awadalla, Ahmed Awadallah, Ammar~Ahmad Awan, Nguyen Bach, Amit Bahree, Arash Bakhtiari, Jianmin Bao, Harkirat Behl, Alon Benhaim, Misha Bilenko, Johan Bjorck, Sébastien Bubeck, Martin Cai, Qin Cai, Vishrav Chaudhary, Dong Chen, Dongdong Chen, Weizhu Chen, Yen-Chun Chen, Yi-Ling Chen, Hao Cheng, Parul Chopra, Xiyang Dai, Matthew Dixon, Ronen Eldan, Victor Fragoso, Jianfeng Gao, Mei Gao, Min Gao, Amit Garg, Allie~Del Giorno, Abhishek Goswami, Suriya Gunasekar, Emman Haider, Junheng Hao, Russell~J. Hewett, Wenxiang Hu, Jamie Huynh, Dan Iter, Sam~Ade Jacobs, Mojan Javaheripi, Xin Jin, Nikos Karampatziakis, Piero Kauffmann, Mahoud Khademi, Dongwoo Kim, Young~Jin Kim, Lev Kurilenko, James~R. Lee, Yin~Tat Lee, Yuanzhi Li, Yunsheng Li, Chen Liang, Lars Liden, Xihui Lin, Zeqi Lin, Ce Liu, Liyuan Liu, Mengchen Liu, Weishung Liu, Xiaodong Liu, Chong Luo, Piyush Madan, Ali Mahmoudzadeh, David Majercak, Matt Mazzola, Caio César~Teodoro Mendes, Arindam Mitra, Hardik Modi, Anh Nguyen,
  Brandon Norick, Barun Patra, Daniel Perez-Becker, Thomas Portet, Reid Pryzant, Heyang Qin, Marko Radmilac, Liliang Ren, Gustavo de Rosa, Corby Rosset, Sambudha Roy, Olatunji Ruwase, Olli Saarikivi, Amin Saied, Adil Salim, Michael Santacroce, Shital Shah, Ning Shang, Hiteshi Sharma, Yelong Shen, Swadheen Shukla, Xia Song, Masahiro Tanaka, Andrea Tupini, Praneetha Vaddamanu, Chunyu Wang, Guanhua Wang, Lijuan Wang, Shuohang Wang, Xin Wang, Yu Wang, Rachel Ward, Wen Wen, Philipp Witte, Haiping Wu, Xiaoxia Wu, Michael Wyatt, Bin Xiao, Can Xu, Jiahang Xu, Weijian Xu, Jilong Xue, Sonali Yadav, Fan Yang, Jianwei Yang, Yifan Yang, Ziyi Yang, Donghan Yu, Lu Yuan, Chenruidong Zhang, Cyril Zhang, Jianwen Zhang, Li~Lyna Zhang, Yi Zhang, Yue Zhang, Yunan Zhang, and Xiren Zhou.
\newblock Phi-3 technical report: A highly capable language model locally on your phone, 2024.

\bibitem[Agrawal et~al.(2024)Agrawal, Jalal, and Sharma]{DBLP:journals/ci/AgrawalJS24a}
Mayank Agrawal, Anand~Singh Jalal, and Himanshu Sharma.
\newblock Enhancing scene-text visual question answering with relational reasoning, attention and dynamic vocabulary integration.
\newblock \emph{Comput. Intell.}, 40\penalty0 (1), 2024.

\bibitem[Alayrac et~al.(2022)Alayrac, Donahue, Luc, Miech, Barr, Hasson, Lenc, Mensch, Millican, Reynolds, Ring, Rutherford, Cabi, Han, Gong, Samangooei, Monteiro, Menick, Borgeaud, Brock, Nematzadeh, Sharifzadeh, Binkowski, Barreira, Vinyals, Zisserman, and Simonyan]{AlayracDLMBHLMM22}
Jean{-}Baptiste Alayrac, Jeff Donahue, Pauline Luc, Antoine Miech, Iain Barr, Yana Hasson, Karel Lenc, Arthur Mensch, Katherine Millican, Malcolm Reynolds, Roman Ring, Eliza Rutherford, Serkan Cabi, Tengda Han, Zhitao Gong, Sina Samangooei, Marianne Monteiro, Jacob~L. Menick, Sebastian Borgeaud, Andy Brock, Aida Nematzadeh, Sahand Sharifzadeh, Mikolaj Binkowski, Ricardo Barreira, Oriol Vinyals, Andrew Zisserman, and Kar{\'{e}}n Simonyan.
\newblock Flamingo: a visual language model for few-shot learning.
\newblock In \emph{Advances in Neural Information Processing Systems 35: Annual Conference on Neural Information Processing Systems 2022, NeurIPS 2022, New Orleans, LA, USA, November 28 - December 9, 2022}, 2022.

\bibitem[Anthropic(2024)]{claude}
Anthropic.
\newblock Claude 3.5 sonnet, 2024.

\bibitem[BAAI(2024)]{InfinityInstruct2024}
BAAI.
\newblock Infinity instruct.
\newblock \emph{arXiv preprint arXiv:2406}, 2024.

\bibitem[Bai et~al.(2023{\natexlab{a}})Bai, Bai, Chu, Cui, Dang, Deng, Fan, Ge, Han, Huang, Hui, Ji, Li, Lin, Lin, Liu, Liu, Lu, Lu, Ma, Men, Ren, Ren, Tan, Tan, Tu, Wang, Wang, Wang, Wu, Xu, Xu, Yang, Yang, Yang, Yang, Yao, Yu, Yuan, Yuan, Zhang, Zhang, Zhang, Zhang, Zhou, Zhou, Zhou, and Zhu]{qwen}
Jinze Bai, Shuai Bai, Yunfei Chu, Zeyu Cui, Kai Dang, Xiaodong Deng, Yang Fan, Wenbin Ge, Yu Han, Fei Huang, Binyuan Hui, Luo Ji, Mei Li, Junyang Lin, Runji Lin, Dayiheng Liu, Gao Liu, Chengqiang Lu, Keming Lu, Jianxin Ma, Rui Men, Xingzhang Ren, Xuancheng Ren, Chuanqi Tan, Sinan Tan, Jianhong Tu, Peng Wang, Shijie Wang, Wei Wang, Shengguang Wu, Benfeng Xu, Jin Xu, An Yang, Hao Yang, Jian Yang, Shusheng Yang, Yang Yao, Bowen Yu, Hongyi Yuan, Zheng Yuan, Jianwei Zhang, Xingxuan Zhang, Yichang Zhang, Zhenru Zhang, Chang Zhou, Jingren Zhou, Xiaohuan Zhou, and Tianhang Zhu.
\newblock Qwen technical report.
\newblock \emph{arXiv preprint arXiv:2309.16609}, 2023{\natexlab{a}}.

\bibitem[Bai et~al.(2023{\natexlab{b}})Bai, Bai, Yang, Wang, Tan, Wang, Lin, Zhou, and Zhou]{abs-2308-12966}
Jinze Bai, Shuai Bai, Shusheng Yang, Shijie Wang, Sinan Tan, Peng Wang, Junyang Lin, Chang Zhou, and Jingren Zhou.
\newblock Qwen-vl: {A} frontier large vision-language model with versatile abilities.
\newblock \emph{CoRR}, abs/2308.12966, 2023{\natexlab{b}}.

\bibitem[Beyer* et~al.(2024)Beyer*, Steiner*, Pinto*, Kolesnikov*, Wang*, Salz, Neumann, Alabdulmohsin, Tschannen, Bugliarello, Unterthiner, Keysers, Koppula, Liu, Grycner, Gritsenko, Houlsby, Kumar, Rong, Eisenschlos, Kabra, Bauer, Bošnjak, Chen, Minderer, Voigtlaender, Bica, Balazevic, Puigcerver, Papalampidi, Henaff, Xiong, Soricut, Harmsen, and Zhai*]{beyer2024paligemma}
Lucas Beyer*, Andreas Steiner*, André~Susano Pinto*, Alexander Kolesnikov*, Xiao Wang*, Daniel Salz, Maxim Neumann, Ibrahim Alabdulmohsin, Michael Tschannen, Emanuele Bugliarello, Thomas Unterthiner, Daniel Keysers, Skanda Koppula, Fangyu Liu, Adam Grycner, Alexey Gritsenko, Neil Houlsby, Manoj Kumar, Keran Rong, Julian Eisenschlos, Rishabh Kabra, Matthias Bauer, Matko Bošnjak, Xi Chen, Matthias Minderer, Paul Voigtlaender, Ioana Bica, Ivana Balazevic, Joan Puigcerver, Pinelopi Papalampidi, Olivier Henaff, Xi Xiong, Radu Soricut, Jeremiah Harmsen, and Xiaohua Zhai*.
\newblock {PaliGemma: A versatile 3B VLM for transfer}.
\newblock \emph{arXiv preprint arXiv:2407.07726}, 2024.

\bibitem[Carter(2024)]{textocr-gpt4v}
Jimmy Carter.
\newblock Textocr-gpt4v.
\newblock \url{https://huggingface.co/datasets/jimmycarter/textocr-gpt4v}, 2024.

\bibitem[Chen et~al.(2024{\natexlab{a}})Chen, Chen, Zhang, Chen, Wu, Zhang, Chen, Li, Wan, and Wang]{chen2024allava}
Guiming~Hardy Chen, Shunian Chen, Ruifei Zhang, Junying Chen, Xiangbo Wu, Zhiyi Zhang, Zhihong Chen, Jianquan Li, Xiang Wan, and Benyou Wang.
\newblock Allava: Harnessing gpt4v-synthesized data for a lite vision-language model, 2024{\natexlab{a}}.

\bibitem[Chen et~al.(2021)Chen, Tang, Qin, Liang, Liu, Xing, and Lin]{DBLP:conf/acl/ChenTQLLXL21}
Jiaqi Chen, Jianheng Tang, Jinghui Qin, Xiaodan Liang, Lingbo Liu, Eric~P. Xing, and Liang Lin.
\newblock Geoqa: {A} geometric question answering benchmark towards multimodal numerical reasoning.
\newblock In \emph{Findings of the Association for Computational Linguistics: {ACL/IJCNLP} 2021, Online Event, August 1-6, 2021}, pages 513--523. Association for Computational Linguistics, 2021.

\bibitem[Chen et~al.(2023)Chen, Li, Dong, Zhang, He, Wang, Zhao, and Lin]{chen2023sharegpt4v}
Lin Chen, Jisong Li, Xiaoyi Dong, Pan Zhang, Conghui He, Jiaqi Wang, Feng Zhao, and Dahua Lin.
\newblock Sharegpt4v: Improving large multi-modal models with better captions.
\newblock \emph{arXiv preprint arXiv:2311.12793}, 2023.

\bibitem[Chen et~al.(2024{\natexlab{b}})Chen, Li, Dong, Zhang, Zang, Chen, Duan, Wang, Qiao, Lin, and Zhao]{DBLP:journals/corr/abs-2403-20330}
Lin Chen, Jinsong Li, Xiaoyi Dong, Pan Zhang, Yuhang Zang, Zehui Chen, Haodong Duan, Jiaqi Wang, Yu Qiao, Dahua Lin, and Feng Zhao.
\newblock Are we on the right way for evaluating large vision-language models?
\newblock \emph{CoRR}, abs/2403.20330, 2024{\natexlab{b}}.

\bibitem[Chen et~al.(2024{\natexlab{c}})Chen, Wang, Tian, Ye, Gao, Cui, Tong, Hu, Luo, Ma, et~al.]{chen2024far}
Zhe Chen, Weiyun Wang, Hao Tian, Shenglong Ye, Zhangwei Gao, Erfei Cui, Wenwen Tong, Kongzhi Hu, Jiapeng Luo, Zheng Ma, et~al.
\newblock How far are we to gpt-4v? closing the gap to commercial multimodal models with open-source suites.
\newblock \emph{arXiv preprint arXiv:2404.16821}, 2024{\natexlab{c}}.

\bibitem[Cylingo(2024)]{xinyuan}
Cylingo.
\newblock Xinyuan-vl-2b, 2024.

\bibitem[Dai et~al.(2023)Dai, Li, Li, Tiong, Zhao, Wang, Li, Fung, and Hoi]{Dai0LTZW0FH23}
Wenliang Dai, Junnan Li, Dongxu Li, Anthony Meng~Huat Tiong, Junqi Zhao, Weisheng Wang, Boyang Li, Pascale Fung, and Steven C.~H. Hoi.
\newblock Instructblip: Towards general-purpose vision-language models with instruction tuning.
\newblock In \emph{Advances in Neural Information Processing Systems 36: Annual Conference on Neural Information Processing Systems 2023, NeurIPS 2023, New Orleans, LA, USA, December 10 - 16, 2023}, 2023.

\bibitem[Das et~al.(2017)Das, Kottur, Gupta, Singh, Yadav, Moura, Parikh, and Batra]{DBLP:conf/cvpr/DasKGSYMPB17}
Abhishek Das, Satwik Kottur, Khushi Gupta, Avi Singh, Deshraj Yadav, Jos{\'{e}} M.~F. Moura, Devi Parikh, and Dhruv Batra.
\newblock Visual dialog.
\newblock In \emph{2017 {IEEE} Conference on Computer Vision and Pattern Recognition, {CVPR} 2017, Honolulu, HI, USA, July 21-26, 2017}, pages 1080--1089. {IEEE} Computer Society, 2017.

\bibitem[Diao et~al.(2024)Diao, Cui, Li, Wang, Lu, and Wang]{abs-2406-11832}
Haiwen Diao, Yufeng Cui, Xiaotong Li, Yueze Wang, Huchuan Lu, and Xinlong Wang.
\newblock Unveiling encoder-free vision-language models.
\newblock \emph{CoRR}, abs/2406.11832, 2024.

\bibitem[Ding et~al.(2021)Ding, Yang, Hong, Zheng, Zhou, Yin, Lin, Zou, Shao, Yang, and Tang]{DingYHZZYLZSYT21}
Ming Ding, Zhuoyi Yang, Wenyi Hong, Wendi Zheng, Chang Zhou, Da Yin, Junyang Lin, Xu Zou, Zhou Shao, Hongxia Yang, and Jie Tang.
\newblock Cogview: Mastering text-to-image generation via transformers.
\newblock In \emph{Advances in Neural Information Processing Systems 34: Annual Conference on Neural Information Processing Systems 2021, NeurIPS 2021, December 6-14, 2021, virtual}, pages 19822--19835, 2021.

\bibitem[Doan et~al.(2024)Doan, Huynh, Hoang, Pham, Pham, Nguyen, Vo, and Hoang]{Vintern-3B}
Khang~T. Doan, Bao~G. Huynh, Dung~T. Hoang, Thuc~D. Pham, Nhat~H. Pham, Quan T.~M. Nguyen, Bang~Q. Vo, and Suong~N. Hoang.
\newblock Vintern-3b-beta, 2024.

\bibitem[Duan et~al.(2024)Duan, Yang, Qiao, Fang, Chen, Liu, Dong, Zang, Zhang, Wang, Lin, and Chen]{duan2024vlmevalkit}
Haodong Duan, Junming Yang, Yuxuan Qiao, Xinyu Fang, Lin Chen, Yuan Liu, Xiaoyi Dong, Yuhang Zang, Pan Zhang, Jiaqi Wang, Dahua Lin, and Kai Chen.
\newblock Vlmevalkit: An open-source toolkit for evaluating large multi-modality models, 2024.

\bibitem[Fu et~al.(2023)Fu, Chen, Shen, Qin, Zhang, Lin, Qiu, Lin, Yang, Zheng, Li, Sun, and Ji]{DBLP:journals/corr/abs-2306-13394}
Chaoyou Fu, Peixian Chen, Yunhang Shen, Yulei Qin, Mengdan Zhang, Xu Lin, Zhenyu Qiu, Wei Lin, Jinrui Yang, Xiawu Zheng, Ke Li, Xing Sun, and Rongrong Ji.
\newblock {MME:} {A} comprehensive evaluation benchmark for multimodal large language models.
\newblock \emph{CoRR}, abs/2306.13394, 2023.

\bibitem[Galib et~al.(2024)Galib, Wang, Xu, Pfeiffer, Chesler, Landry, and Ambati]{galib2024h2ovlmississippivisionlanguagemodels}
Shaikat Galib, Shanshan Wang, Guanshuo Xu, Pascal Pfeiffer, Ryan Chesler, Mark Landry, and Sri~Satish Ambati.
\newblock H2ovl-mississippi vision language models technical report, 2024.

\bibitem[Guan et~al.(2024)Guan, Liu, Wu, Xian, Li, Liu, Wang, Chen, Huang, Yacoob, Manocha, and Zhou]{DBLP:conf/cvpr/GuanLWXLL0CHYM024}
Tianrui Guan, Fuxiao Liu, Xiyang Wu, Ruiqi Xian, Zongxia Li, Xiaoyu Liu, Xijun Wang, Lichang Chen, Furong Huang, Yaser Yacoob, Dinesh Manocha, and Tianyi Zhou.
\newblock Hallusionbench: An advanced diagnostic suite for entangled language hallucination and visual illusion in large vision-language models.
\newblock In \emph{{CVPR}}, pages 14375--14385. {IEEE}, 2024.

\bibitem[Gupta et~al.(2019)Gupta, Doll{\'{a}}r, and Girshick]{DBLP:conf/cvpr/GuptaDG19}
Agrim Gupta, Piotr Doll{\'{a}}r, and Ross~B. Girshick.
\newblock {LVIS:} {A} dataset for large vocabulary instance segmentation.
\newblock In \emph{{IEEE} Conference on Computer Vision and Pattern Recognition, {CVPR} 2019, Long Beach, CA, USA, June 16-20, 2019}, pages 5356--5364. Computer Vision Foundation / {IEEE}, 2019.

\bibitem[Havard et~al.(2017)Havard, Besacier, and Rosec]{DBLP:journals/corr/HavardBR17}
William Havard, Laurent Besacier, and Olivier Rosec.
\newblock {SPEECH-COCO:} 600k visually grounded spoken captions aligned to {MSCOCO} data set.
\newblock \emph{CoRR}, abs/1707.08435, 2017.

\bibitem[Hendrycks and Gimpel(2023)]{hendrycks2023gaussianerrorlinearunits}
Dan Hendrycks and Kevin Gimpel.
\newblock Gaussian error linear units (gelus), 2023.

\bibitem[Huang et~al.(2024)Huang, Liu, Liang, Jin, and Bai]{huang2024mini}
Mingxin Huang, Yuliang Liu, Dingkang Liang, Lianwen Jin, and Xiang Bai.
\newblock Mini-monkey: Multi-scale adaptive cropping for multimodal large language models.
\newblock \emph{arXiv preprint arXiv:2408.02034}, 2024.

\bibitem[Huang et~al.(2023)Huang, Huang, Zhang, Tian, Feng, Zhang, Xie, Li, and Zhang]{huang2023open}
Xinyu Huang, Yi-Jie Huang, Youcai Zhang, Weiwei Tian, Rui Feng, Yuejie Zhang, Yanchun Xie, Yaqian Li, and Lei Zhang.
\newblock Open-set image tagging with multi-grained text supervision.
\newblock \emph{arXiv e-prints}, pages arXiv--2310, 2023.

\bibitem[Kembhavi et~al.(2016)Kembhavi, Salvato, Kolve, Seo, Hajishirzi, and Farhadi]{DBLP:conf/eccv/KembhaviSKSHF16}
Aniruddha Kembhavi, Mike Salvato, Eric Kolve, Min~Joon Seo, Hannaneh Hajishirzi, and Ali Farhadi.
\newblock A diagram is worth a dozen images.
\newblock In \emph{{ECCV} {(4)}}, pages 235--251. Springer, 2016.

\bibitem[Lauren{\c{c}}on et~al.(2024)Lauren{\c{c}}on, Marafioti, Sanh, and Tronchon]{DBLP:journals/corr/abs-2408-12637}
Hugo Lauren{\c{c}}on, Andr{\'{e}}s Marafioti, Victor Sanh, and L{\'{e}}o Tronchon.
\newblock Building and better understanding vision-language models: insights and future directions.
\newblock \emph{CoRR}, abs/2408.12637, 2024.

\bibitem[Li et~al.(2024{\natexlab{a}})Li, Ge, Chen, Ge, Zhang, and Shan]{DBLP:journals/corr/abs-2404-16790}
Bohao Li, Yuying Ge, Yi Chen, Yixiao Ge, Ruimao Zhang, and Ying Shan.
\newblock Seed-bench-2-plus: Benchmarking multimodal large language models with text-rich visual comprehension.
\newblock \emph{CoRR}, abs/2404.16790, 2024{\natexlab{a}}.

\bibitem[Li et~al.(2024{\natexlab{b}})Li, Zhang, Guo, Zhang, Li, Zhang, Zhang, Li, Liu, and Li]{DBLP:journals/corr/abs-2408-03326}
Bo Li, Yuanhan Zhang, Dong Guo, Renrui Zhang, Feng Li, Hao Zhang, Kaichen Zhang, Yanwei Li, Ziwei Liu, and Chunyuan Li.
\newblock Llava-onevision: Easy visual task transfer.
\newblock \emph{CoRR}, abs/2408.03326, 2024{\natexlab{b}}.

\bibitem[Li et~al.(2024{\natexlab{c}})Li, Zhang, Zhang, Zhang, Li, Li, Ma, and Li]{li2024llava}
Feng Li, Renrui Zhang, Hao Zhang, Yuanhan Zhang, Bo Li, Wei Li, Zejun Ma, and Chunyuan Li.
\newblock Llava-next-interleave: Tackling multi-image, video, and 3d in large multimodal models.
\newblock \emph{arXiv preprint arXiv:2407.07895}, 2024{\natexlab{c}}.

\bibitem[Li et~al.(2023{\natexlab{a}})Li, Li, Savarese, and Hoi]{0008LSH23}
Junnan Li, Dongxu Li, Silvio Savarese, and Steven C.~H. Hoi.
\newblock {BLIP-2:} bootstrapping language-image pre-training with frozen image encoders and large language models.
\newblock In \emph{International Conference on Machine Learning, {ICML} 2023, 23-29 July 2023, Honolulu, Hawaii, {USA}}, pages 19730--19742. {PMLR}, 2023{\natexlab{a}}.

\bibitem[Li et~al.(2024{\natexlab{d}})Li, Zhang, Diao, Wang, Wang, and Duan]{abs-2407-08303}
Xiaotong Li, Fan Zhang, Haiwen Diao, Yueze Wang, Xinlong Wang, and Ling{-}Yu Duan.
\newblock Densefusion-1m: Merging vision experts for comprehensive multimodal perception.
\newblock \emph{CoRR}, abs/2407.08303, 2024{\natexlab{d}}.

\bibitem[Li et~al.(2023{\natexlab{b}})Li, Du, Zhou, Wang, Zhao, and Wen]{DBLP:conf/emnlp/LiDZWZW23}
Yifan Li, Yifan Du, Kun Zhou, Jinpeng Wang, Wayne~Xin Zhao, and Ji{-}Rong Wen.
\newblock Evaluating object hallucination in large vision-language models.
\newblock In \emph{{EMNLP}}, pages 292--305. Association for Computational Linguistics, 2023{\natexlab{b}}.

\bibitem[Liu et~al.(2024{\natexlab{a}})Liu, Wang, Yao, Chen, Song, Cho, Yacoob, and Yu]{DBLP:conf/naacl/LiuWYCSCYY24}
Fuxiao Liu, Xiaoyang Wang, Wenlin Yao, Jianshu Chen, Kaiqiang Song, Sangwoo Cho, Yaser Yacoob, and Dong Yu.
\newblock {MMC:} advancing multimodal chart understanding with large-scale instruction tuning.
\newblock In \emph{Proceedings of the 2024 Conference of the North American Chapter of the Association for Computational Linguistics: Human Language Technologies (Volume 1: Long Papers), {NAACL} 2024, Mexico City, Mexico, June 16-21, 2024}, pages 1287--1310. Association for Computational Linguistics, 2024{\natexlab{a}}.

\bibitem[Liu et~al.(2023{\natexlab{a}})Liu, Li, Wu, and Lee]{DBLP:conf/nips/LiuLWL23a}
Haotian Liu, Chunyuan Li, Qingyang Wu, and Yong~Jae Lee.
\newblock Visual instruction tuning.
\newblock In \emph{NeurIPS}, 2023{\natexlab{a}}.

\bibitem[Liu et~al.(2023{\natexlab{b}})Liu, Li, Wu, and Lee]{LiuLWL23a}
Haotian Liu, Chunyuan Li, Qingyang Wu, and Yong~Jae Lee.
\newblock Visual instruction tuning.
\newblock In \emph{Advances in Neural Information Processing Systems 36: Annual Conference on Neural Information Processing Systems 2023, NeurIPS 2023, New Orleans, LA, USA, December 10 - 16, 2023}, 2023{\natexlab{b}}.

\bibitem[Liu et~al.(2024{\natexlab{b}})Liu, Li, Li, and Lee]{liu2024improved}
Haotian Liu, Chunyuan Li, Yuheng Li, and Yong~Jae Lee.
\newblock Improved baselines with visual instruction tuning.
\newblock In \emph{{IEEE/CVF} Conference on Computer Vision and Pattern Recognition, {CVPR} 2024, Seattle, WA, USA, June 16-22, 2024}, pages 26286--26296. {IEEE}, 2024{\natexlab{b}}.

\bibitem[Liu et~al.(2023{\natexlab{c}})Liu, Duan, Zhang, Li, Zhang, Zhao, Yuan, Wang, He, Liu, Chen, and Lin]{DBLP:journals/corr/abs-2307-06281}
Yuan Liu, Haodong Duan, Yuanhan Zhang, Bo Li, Songyang Zhang, Wangbo Zhao, Yike Yuan, Jiaqi Wang, Conghui He, Ziwei Liu, Kai Chen, and Dahua Lin.
\newblock Mmbench: Is your multi-modal model an all-around player?
\newblock \emph{CoRR}, abs/2307.06281, 2023{\natexlab{c}}.

\bibitem[Liu et~al.(2023{\natexlab{d}})Liu, Li, Li, Yu, Huang, Peng, Liu, Chen, Li, Jin, and Bai]{DBLP:journals/corr/abs-2305-07895}
Yuliang Liu, Zhang Li, Hongliang Li, Wenwen Yu, Mingxin Huang, Dezhi Peng, Mingyu Liu, Mingrui Chen, Chunyuan Li, Lianwen Jin, and Xiang Bai.
\newblock On the hidden mystery of {OCR} in large multimodal models.
\newblock \emph{CoRR}, abs/2305.07895, 2023{\natexlab{d}}.

\bibitem[Lu et~al.(2024{\natexlab{a}})Lu, Liu, Zhang, Wang, Dong, Liu, Sun, Ren, Li, Yang, Sun, Deng, Xu, Xie, and Ruan]{lu2024deepseekvl}
Haoyu Lu, Wen Liu, Bo Zhang, Bingxuan Wang, Kai Dong, Bo Liu, Jingxiang Sun, Tongzheng Ren, Zhuoshu Li, Hao Yang, Yaofeng Sun, Chengqi Deng, Hanwei Xu, Zhenda Xie, and Chong Ruan.
\newblock Deepseek-vl: Towards real-world vision-language understanding, 2024{\natexlab{a}}.

\bibitem[Lu et~al.(2022)Lu, Mishra, Xia, Qiu, Chang, Zhu, Tafjord, Clark, and Kalyan]{DBLP:conf/nips/LuMX0CZTCK22}
Pan Lu, Swaroop Mishra, Tanglin Xia, Liang Qiu, Kai{-}Wei Chang, Song{-}Chun Zhu, Oyvind Tafjord, Peter Clark, and Ashwin Kalyan.
\newblock Learn to explain: Multimodal reasoning via thought chains for science question answering.
\newblock In \emph{NeurIPS}, 2022.

\bibitem[Lu et~al.(2024{\natexlab{b}})Lu, Bansal, Xia, Liu, Li, Hajishirzi, Cheng, Chang, Galley, and Gao]{DBLP:conf/iclr/LuBX0LH0CG024}
Pan Lu, Hritik Bansal, Tony Xia, Jiacheng Liu, Chunyuan Li, Hannaneh Hajishirzi, Hao Cheng, Kai{-}Wei Chang, Michel Galley, and Jianfeng Gao.
\newblock Mathvista: Evaluating mathematical reasoning of foundation models in visual contexts.
\newblock In \emph{The Twelfth International Conference on Learning Representations, {ICLR} 2024, Vienna, Austria, May 7-11, 2024}. OpenReview.net, 2024{\natexlab{b}}.

\bibitem[Luo et~al.(2024)Luo, Zhang, Chen, Lin, Liu, Wu, Yang, Wang, Zeng, Gao, Shen, Li, Xia, Huang, Song, and Li]{DBLP:journals/corr/abs-2409-05840}
Run Luo, Haonan Zhang, Longze Chen, Ting{-}En Lin, Xiong Liu, Yuchuan Wu, Min Yang, Minzheng Wang, Pengpeng Zeng, Lianli Gao, Heng~Tao Shen, Yunshui Li, Xiaobo Xia, Fei Huang, Jingkuan Song, and Yongbin Li.
\newblock Mmevol: Empowering multimodal large language models with evol-instruct.
\newblock \emph{CoRR}, abs/2409.05840, 2024.

\bibitem[Masry et~al.(2022)Masry, Long, Tan, Joty, and Hoque]{DBLP:journals/corr/abs-2203-10244}
Ahmed Masry, Do~Xuan Long, Jia~Qing Tan, Shafiq~R. Joty, and Enamul Hoque.
\newblock Chartqa: {A} benchmark for question answering about charts with visual and logical reasoning.
\newblock \emph{CoRR}, abs/2203.10244, 2022.

\bibitem[Mathew et~al.(2021)Mathew, Karatzas, and Jawahar]{DBLP:conf/wacv/MathewKJ21}
Minesh Mathew, Dimosthenis Karatzas, and C.~V. Jawahar.
\newblock Docvqa: {A} dataset for {VQA} on document images.
\newblock In \emph{{IEEE} Winter Conference on Applications of Computer Vision, {WACV} 2021, Waikoloa, HI, USA, January 3-8, 2021}, pages 2199--2208. {IEEE}, 2021.

\bibitem[Mishra et~al.(2019)Mishra, Shekhar, Singh, and Chakraborty]{DBLP:conf/icdar/0001SSC19}
Anand Mishra, Shashank Shekhar, Ajeet~Kumar Singh, and Anirban Chakraborty.
\newblock {OCR-VQA:} visual question answering by reading text in images.
\newblock In \emph{{ICDAR}}, pages 947--952. {IEEE}, 2019.

\bibitem[OpenAI(2024)]{gpt4v_system_card}
OpenAI.
\newblock Gpt-4v system card, 2024.

\bibitem[Peebles and Xie(2023)]{PeeblesX23}
William Peebles and Saining Xie.
\newblock Scalable diffusion models with transformers.
\newblock In \emph{{IEEE/CVF} International Conference on Computer Vision, {ICCV} 2023, Paris, France, October 1-6, 2023}, pages 4172--4182. {IEEE}, 2023.

\bibitem[Radford et~al.(2021)Radford, Kim, Hallacy, Ramesh, Goh, Agarwal, Sastry, Askell, Mishkin, Clark, Krueger, and Sutskever]{RadfordKHRGASAM21}
Alec Radford, Jong~Wook Kim, Chris Hallacy, Aditya Ramesh, Gabriel Goh, Sandhini Agarwal, Girish Sastry, Amanda Askell, Pamela Mishkin, Jack Clark, Gretchen Krueger, and Ilya Sutskever.
\newblock Learning transferable visual models from natural language supervision.
\newblock In \emph{Proceedings of the 38th International Conference on Machine Learning, {ICML} 2021, 18-24 July 2021, Virtual Event}, pages 8748--8763. {PMLR}, 2021.

\bibitem[Ramesh et~al.(2021)Ramesh, Pavlov, Goh, Gray, Voss, Radford, Chen, and Sutskever]{RameshPGGVRCS21}
Aditya Ramesh, Mikhail Pavlov, Gabriel Goh, Scott Gray, Chelsea Voss, Alec Radford, Mark Chen, and Ilya Sutskever.
\newblock Zero-shot text-to-image generation.
\newblock In \emph{Proceedings of the 38th International Conference on Machine Learning, {ICML} 2021, 18-24 July 2021, Virtual Event}, pages 8821--8831. {PMLR}, 2021.

\bibitem[Shi et~al.(2024)Shi, Hu, Bin, Liu, Yang, Ng, Bing, and Lee]{DBLP:journals/corr/abs-2406-17294}
Wenhao Shi, Zhiqiang Hu, Yi Bin, Junhua Liu, Yang Yang, See{-}Kiong Ng, Lidong Bing, and Roy~Ka{-}Wei Lee.
\newblock Math-llava: Bootstrapping mathematical reasoning for multimodal large language models.
\newblock \emph{CoRR}, abs/2406.17294, 2024.

\bibitem[Shi et~al.(2020)Shi, Zhou, Qiu, and Zhu]{shi-etal-2020-improving}
Zhan Shi, Xu Zhou, Xipeng Qiu, and Xiaodan Zhu.
\newblock Improving image captioning with better use of caption.
\newblock In \emph{Proceedings of the 58th Annual Meeting of the Association for Computational Linguistics}, pages 7454--7464, Online, 2020. Association for Computational Linguistics.

\bibitem[Singh et~al.(2019)Singh, Natarajan, Shah, Jiang, Chen, Batra, Parikh, and Rohrbach]{DBLP:conf/cvpr/SinghNSJCBPR19}
Amanpreet Singh, Vivek Natarajan, Meet Shah, Yu Jiang, Xinlei Chen, Dhruv Batra, Devi Parikh, and Marcus Rohrbach.
\newblock Towards {VQA} models that can read.
\newblock In \emph{{CVPR}}, pages 8317--8326. Computer Vision Foundation / {IEEE}, 2019.

\bibitem[Sun et~al.(2024{\natexlab{a}})Sun, Cui, Zhang, Zhang, Yu, Wang, Rao, Liu, Huang, and Wang]{SunCZZYWRL0W24}
Quan Sun, Yufeng Cui, Xiaosong Zhang, Fan Zhang, Qiying Yu, Yueze Wang, Yongming Rao, Jingjing Liu, Tiejun Huang, and Xinlong Wang.
\newblock Generative multimodal models are in-context learners.
\newblock In \emph{{IEEE/CVF} Conference on Computer Vision and Pattern Recognition, {CVPR} 2024, Seattle, WA, USA, June 16-22, 2024}, pages 14398--14409. {IEEE}, 2024{\natexlab{a}}.

\bibitem[Sun et~al.(2024{\natexlab{b}})Sun, Yu, Cui, Zhang, Zhang, Wang, Gao, Liu, Huang, and Wang]{SunYCZZWGL0W24}
Quan Sun, Qiying Yu, Yufeng Cui, Fan Zhang, Xiaosong Zhang, Yueze Wang, Hongcheng Gao, Jingjing Liu, Tiejun Huang, and Xinlong Wang.
\newblock Emu: Generative pretraining in multimodality.
\newblock In \emph{The Twelfth International Conference on Learning Representations, {ICLR} 2024, Vienna, Austria, May 7-11, 2024}. OpenReview.net, 2024{\natexlab{b}}.

\bibitem[Tong et~al.(2024)Tong, Brown, Wu, Woo, Middepogu, Akula, Yang, Yang, Iyer, Pan, Wang, Fergus, LeCun, and Xie]{DBLP:journals/corr/abs-2406-16860}
Shengbang Tong, Ellis Brown, Penghao Wu, Sanghyun Woo, Manoj Middepogu, Sai~Charitha Akula, Jihan Yang, Shusheng Yang, Adithya Iyer, Xichen Pan, Austin Wang, Rob Fergus, Yann LeCun, and Saining Xie.
\newblock Cambrian-1: {A} fully open, vision-centric exploration of multimodal llms.
\newblock \emph{CoRR}, abs/2406.16860, 2024.

\bibitem[Wang et~al.(2023{\natexlab{a}})Wang, Meng, Weng, He, Wu, and Jiang]{abs-2311-07574}
Junke Wang, Lingchen Meng, Zejia Weng, Bo He, Zuxuan Wu, and Yu{-}Gang Jiang.
\newblock To see is to believe: Prompting {GPT-4V} for better visual instruction tuning.
\newblock \emph{CoRR}, abs/2311.07574, 2023{\natexlab{a}}.

\bibitem[Wang et~al.(2024{\natexlab{a}})Wang, Pan, Shi, Lu, Zhan, and Li]{DBLP:journals/corr/abs-2402-14804}
Ke Wang, Junting Pan, Weikang Shi, Zimu Lu, Mingjie Zhan, and Hongsheng Li.
\newblock Measuring multimodal mathematical reasoning with math-vision dataset.
\newblock \emph{CoRR}, abs/2402.14804, 2024{\natexlab{a}}.

\bibitem[Wang et~al.(2024{\natexlab{b}})Wang, Bai, Tan, Wang, Fan, Bai, Chen, Liu, Wang, Ge, Fan, Dang, Du, Ren, Men, Liu, Zhou, Zhou, and Lin]{DBLP:journals/corr/abs-2409-12191}
Peng Wang, Shuai Bai, Sinan Tan, Shijie Wang, Zhihao Fan, Jinze Bai, Keqin Chen, Xuejing Liu, Jialin Wang, Wenbin Ge, Yang Fan, Kai Dang, Mengfei Du, Xuancheng Ren, Rui Men, Dayiheng Liu, Chang Zhou, Jingren Zhou, and Junyang Lin.
\newblock Qwen2-vl: Enhancing vision-language model's perception of the world at any resolution.
\newblock \emph{CoRR}, abs/2409.12191, 2024{\natexlab{b}}.

\bibitem[Wang et~al.(2023{\natexlab{b}})Wang, Lv, Yu, Hong, Qi, Wang, Ji, Yang, Zhao, Song, Xu, Xu, Li, Dong, Ding, and Tang]{abs-2311-03079}
Weihan Wang, Qingsong Lv, Wenmeng Yu, Wenyi Hong, Ji Qi, Yan Wang, Junhui Ji, Zhuoyi Yang, Lei Zhao, Xixuan Song, Jiazheng Xu, Bin Xu, Juanzi Li, Yuxiao Dong, Ming Ding, and Jie Tang.
\newblock Cogvlm: Visual expert for pretrained language models.
\newblock \emph{CoRR}, abs/2311.03079, 2023{\natexlab{b}}.

\bibitem[Wang et~al.(2024{\natexlab{c}})Wang, Zhang, Luo, Sun, Cui, Wang, Zhang, Wang, Li, Yu, Zhao, Ao, Min, Li, Wu, Zhao, Zhang, Wang, Liu, He, Yang, Liu, Lin, Huang, and Wang]{wang2024emu3nexttokenpredictionneed}
Xinlong Wang, Xiaosong Zhang, Zhengxiong Luo, Quan Sun, Yufeng Cui, Jinsheng Wang, Fan Zhang, Yueze Wang, Zhen Li, Qiying Yu, Yingli Zhao, Yulong Ao, Xuebin Min, Tao Li, Boya Wu, Bo Zhao, Bowen Zhang, Liangdong Wang, Guang Liu, Zheqi He, Xi Yang, Jingjing Liu, Yonghua Lin, Tiejun Huang, and Zhongyuan Wang.
\newblock Emu3: Next-token prediction is all you need, 2024{\natexlab{c}}.

\bibitem[Wang et~al.(2020)Wang, Zhan, Liu, and Liang]{DBLP:conf/emnlp/WangZ0L20}
Zilong Wang, Mingjie Zhan, Xuebo Liu, and Ding Liang.
\newblock Docstruct: {A} multimodal method to extract hierarchy structure in document for general form understanding.
\newblock In \emph{Findings of the Association for Computational Linguistics: {EMNLP} 2020, Online Event, 16-20 November 2020}, pages 898--908. Association for Computational Linguistics, 2020.

\bibitem[{X.AI}(2024)]{grok2024}
{X.AI}.
\newblock {Grok-1.5 Vision Preview}.
\newblock \url{https://x.ai/blog/grok-1.5v}, 2024.

\bibitem[Xiao et~al.(2024)Xiao, Wu, Xu, Dai, Hu, Lu, Zeng, Liu, and Yuan]{0004WXDHL00Y24}
Bin Xiao, Haiping Wu, Weijian Xu, Xiyang Dai, Houdong Hu, Yumao Lu, Michael Zeng, Ce Liu, and Lu Yuan.
\newblock Florence-2: Advancing a unified representation for a variety of vision tasks.
\newblock In \emph{{IEEE/CVF} Conference on Computer Vision and Pattern Recognition, {CVPR} 2024, Seattle, WA, USA, June 16-22, 2024}, pages 4818--4829. {IEEE}, 2024.

\bibitem[Xie et~al.(2024)Xie, Mao, Bai, Zhang, Wang, Lin, Gu, Chen, Yang, and Shou]{abs-2408-12528}
Jinheng Xie, Weijia Mao, Zechen Bai, David~Junhao Zhang, Weihao Wang, Kevin~Qinghong Lin, Yuchao Gu, Zhijie Chen, Zhenheng Yang, and Mike~Zheng Shou.
\newblock Show-o: One single transformer to unify multimodal understanding and generation.
\newblock \emph{CoRR}, abs/2408.12528, 2024.

\bibitem[Yao et~al.(2024)Yao, Yu, Zhang, Wang, Cui, Zhu, Cai, Li, Zhao, He, Chen, Zhou, Zou, Zhang, Hu, Zheng, Zhou, Cai, Han, Zeng, Li, Liu, and Sun]{yao2024minicpmvgpt4vlevelmllm}
Yuan Yao, Tianyu Yu, Ao Zhang, Chongyi Wang, Junbo Cui, Hongji Zhu, Tianchi Cai, Haoyu Li, Weilin Zhao, Zhihui He, Qianyu Chen, Huarong Zhou, Zhensheng Zou, Haoye Zhang, Shengding Hu, Zhi Zheng, Jie Zhou, Jie Cai, Xu Han, Guoyang Zeng, Dahai Li, Zhiyuan Liu, and Maosong Sun.
\newblock Minicpm-v: A gpt-4v level mllm on your phone.
\newblock \emph{arXiv preprint arXiv:2408.01800}, 2024.

\bibitem[Ye et~al.(2024)Ye, Xu, Ye, Yan, Hu, Liu, Qian, Zhang, and Huang]{DBLP:conf/cvpr/YeXYYHL0Z024}
Qinghao Ye, Haiyang Xu, Jiabo Ye, Ming Yan, Anwen Hu, Haowei Liu, Qi Qian, Ji Zhang, and Fei Huang.
\newblock mplug-owi2: Revolutionizing multi-modal large language model with modality collaboration.
\newblock In \emph{{IEEE/CVF} Conference on Computer Vision and Pattern Recognition, {CVPR} 2024, Seattle, WA, USA, June 16-22, 2024}, pages 13040--13051. {IEEE}, 2024.

\bibitem[Ying et~al.(2024)Ying, Meng, Wang, Li, Lin, Yang, Zhang, Zhang, Lin, Liu, Lei, Lu, Chen, Xu, Zhang, Zhang, Gao, Wang, Qiao, Luo, Zhang, and Shao]{DBLP:conf/icml/YingMWLLYZZLLLL24}
Kaining Ying, Fanqing Meng, Jin Wang, Zhiqian Li, Han Lin, Yue Yang, Hao Zhang, Wenbo Zhang, Yuqi Lin, Shuo Liu, Jiayi Lei, Quanfeng Lu, Runjian Chen, Peng Xu, Renrui Zhang, Haozhe Zhang, Peng Gao, Yali Wang, Yu Qiao, Ping Luo, Kaipeng Zhang, and Wenqi Shao.
\newblock Mmt-bench: {A} comprehensive multimodal benchmark for evaluating large vision-language models towards multitask {AGI}.
\newblock In \emph{{ICML}}. OpenReview.net, 2024.

\bibitem[Yu et~al.(2023)Yu, Yang, Li, Wang, Lin, Liu, Wang, and Wang]{DBLP:journals/corr/abs-2308-02490}
Weihao Yu, Zhengyuan Yang, Linjie Li, Jianfeng Wang, Kevin Lin, Zicheng Liu, Xinchao Wang, and Lijuan Wang.
\newblock Mm-vet: Evaluating large multimodal models for integrated capabilities.
\newblock \emph{CoRR}, abs/2308.02490, 2023.

\bibitem[Yue et~al.(2023)Yue, Ni, Zhang, Zheng, Liu, Zhang, Stevens, Jiang, Ren, Sun, Wei, Yu, Yuan, Sun, Yin, Zheng, Yang, Liu, Huang, Sun, Su, and Chen]{DBLP:journals/corr/abs-2311-16502}
Xiang Yue, Yuansheng Ni, Kai Zhang, Tianyu Zheng, Ruoqi Liu, Ge Zhang, Samuel Stevens, Dongfu Jiang, Weiming Ren, Yuxuan Sun, Cong Wei, Botao Yu, Ruibin Yuan, Renliang Sun, Ming Yin, Boyuan Zheng, Zhenzhu Yang, Yibo Liu, Wenhao Huang, Huan Sun, Yu Su, and Wenhu Chen.
\newblock {MMMU:} {A} massive multi-discipline multimodal understanding and reasoning benchmark for expert {AGI}.
\newblock \emph{CoRR}, abs/2311.16502, 2023.

\bibitem[Zauner(2010)]{Zauner2010ImplementationAB}
Christoph Zauner.
\newblock Implementation and benchmarking of perceptual image hash functions.
\newblock 2010.

\bibitem[Zhai et~al.(2023)Zhai, Mustafa, Kolesnikov, and Beyer]{zhai2023sigmoid}
Xiaohua Zhai, Basil Mustafa, Alexander Kolesnikov, and Lucas Beyer.
\newblock Sigmoid loss for language image pre-training.
\newblock In \emph{Proceedings of the IEEE/CVF International Conference on Computer Vision}, pages 11975--11986, 2023.

\bibitem[Zhang et~al.(2024{\natexlab{a}})Zhang, Jiang, Zhang, Lin, Guo, Qiu, Zhou, Lu, Chang, Gao, and Li]{DBLP:journals/corr/abs-2403-14624}
Renrui Zhang, Dongzhi Jiang, Yichi Zhang, Haokun Lin, Ziyu Guo, Pengshuo Qiu, Aojun Zhou, Pan Lu, Kai{-}Wei Chang, Peng Gao, and Hongsheng Li.
\newblock Mathverse: Does your multi-modal {LLM} truly see the diagrams in visual math problems?
\newblock \emph{CoRR}, abs/2403.14624, 2024{\natexlab{a}}.

\bibitem[Zhang et~al.(2024{\natexlab{b}})Zhang, Wang, Li, Zhang, Taslakian, Rajeswar, Fu, Liu, and Bengio]{DBLP:journals/corr/abs-2406-06462}
Tianyu Zhang, Suyuchen Wang, Lu Li, Ge Zhang, Perouz Taslakian, Sai Rajeswar, Jie Fu, Bang Liu, and Yoshua Bengio.
\newblock {VCR:} visual caption restoration.
\newblock \emph{CoRR}, abs/2406.06462, 2024{\natexlab{b}}.

\bibitem[Zhou et~al.(2024)Zhou, Yu, Babu, Tirumala, Yasunaga, Shamis, Kahn, Ma, Zettlemoyer, and Levy]{abs-2408-11039}
Chunting Zhou, Lili Yu, Arun Babu, Kushal Tirumala, Michihiro Yasunaga, Leonid Shamis, Jacob Kahn, Xuezhe Ma, Luke Zettlemoyer, and Omer Levy.
\newblock Transfusion: Predict the next token and diffuse images with one multi-modal model.
\newblock \emph{CoRR}, abs/2408.11039, 2024.

\bibitem[Zhu et~al.(2024)Zhu, Chen, Shen, Li, and Elhoseiny]{Zhu0SLE24}
Deyao Zhu, Jun Chen, Xiaoqian Shen, Xiang Li, and Mohamed Elhoseiny.
\newblock Minigpt-4: Enhancing vision-language understanding with advanced large language models.
\newblock In \emph{The Twelfth International Conference on Learning Representations, {ICLR} 2024, Vienna, Austria, May 7-11, 2024}. OpenReview.net, 2024.

\end{thebibliography}
}

\clearpage \appendix 

\label{sec:appendix}

\section{Video Understanding}

To enhance Aquila-VL-2B's ability to process multi-image and video data, we extracted a total of 937K multi-image and video samples from the LLaVA-OneVision dataset, and combined them with 1M single-image samples drawn from Stage 4 for further training. The results, as shown in Table~\ref{tab:video}, demonstrate that even prior to incorporating the multi-image and video data, our model already exhibited a solid ability to handle video imagery with satisfactory performance. After introducing the additional multi-image and video data for further training, the model's capacity to process such data was significantly improved.

 
\textbf{}
\begin{table*}[]
\centering
\resizebox{\textwidth}{!}{%
\begin{tabular}{l|cccccc}
\toprule
 Benchmark & MiniCPM-V-2 & InternVL2-2B & Qwen2-VL-2B-Instruct & Aquila-VL-2B & Aquila-VL-2B-video \\ \midrule
 Video-MME(w/o subs) &  38.6 & 45.9  & 55.6 & 48.4 & 51.5 \\ \bottomrule

\end{tabular}%
}
\caption{Performance of Aquila-VL-2B and other models on video benchmarks.}
\label{tab:video}
\end{table*}

\section{Data composition}
\label{appen:data}
\begin{figure*}[t]
    \centering
    \begin{subfigure}[b]{0.66\columnwidth}
        \centering
        \includegraphics[width=\textwidth]{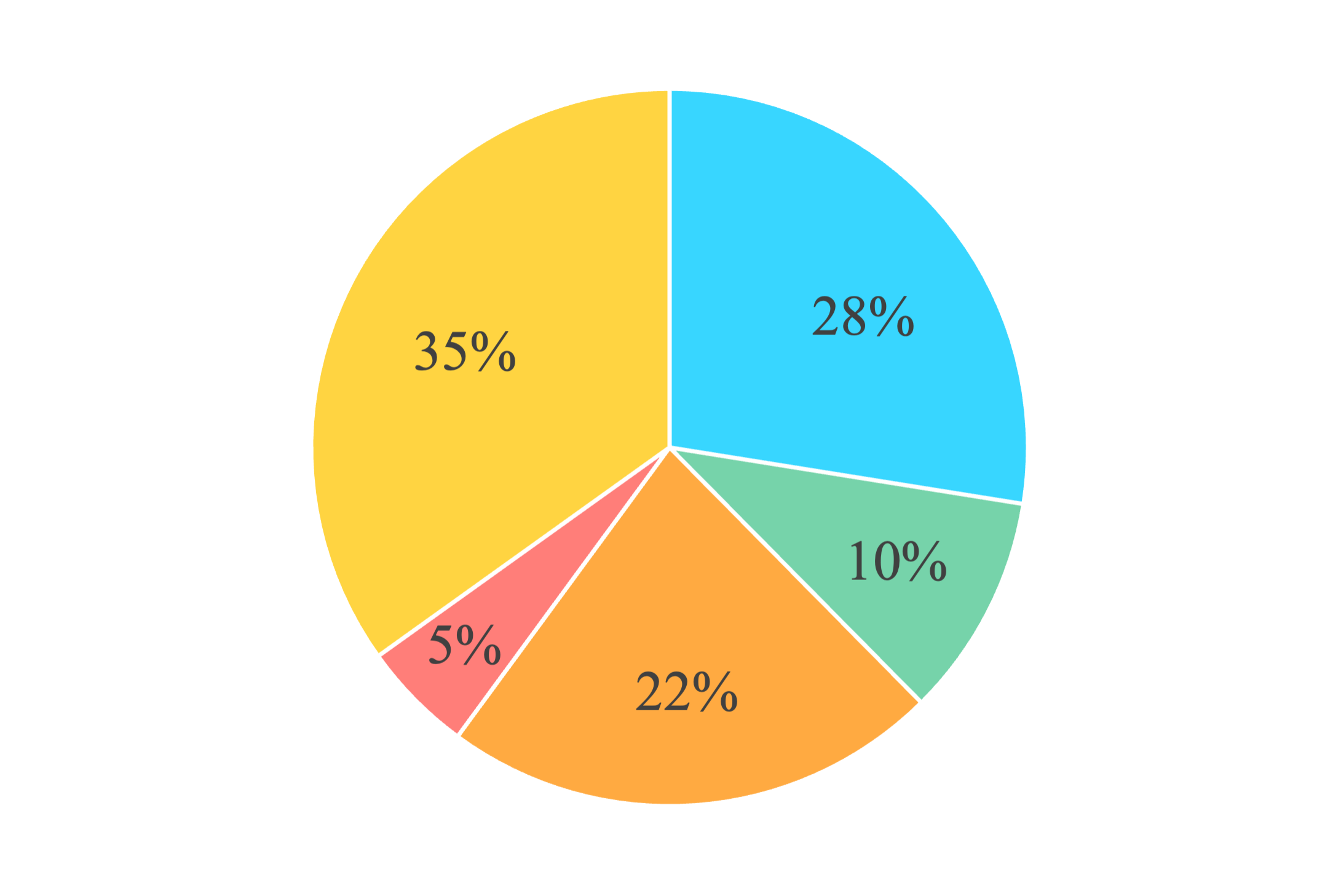}
        \caption{Comprehensive Data}
        \label{fig:stage2_distribution}
    \end{subfigure}%
    \hfill
    \begin{subfigure}[b]{0.66\columnwidth}
        \centering
        \includegraphics[width=\textwidth]{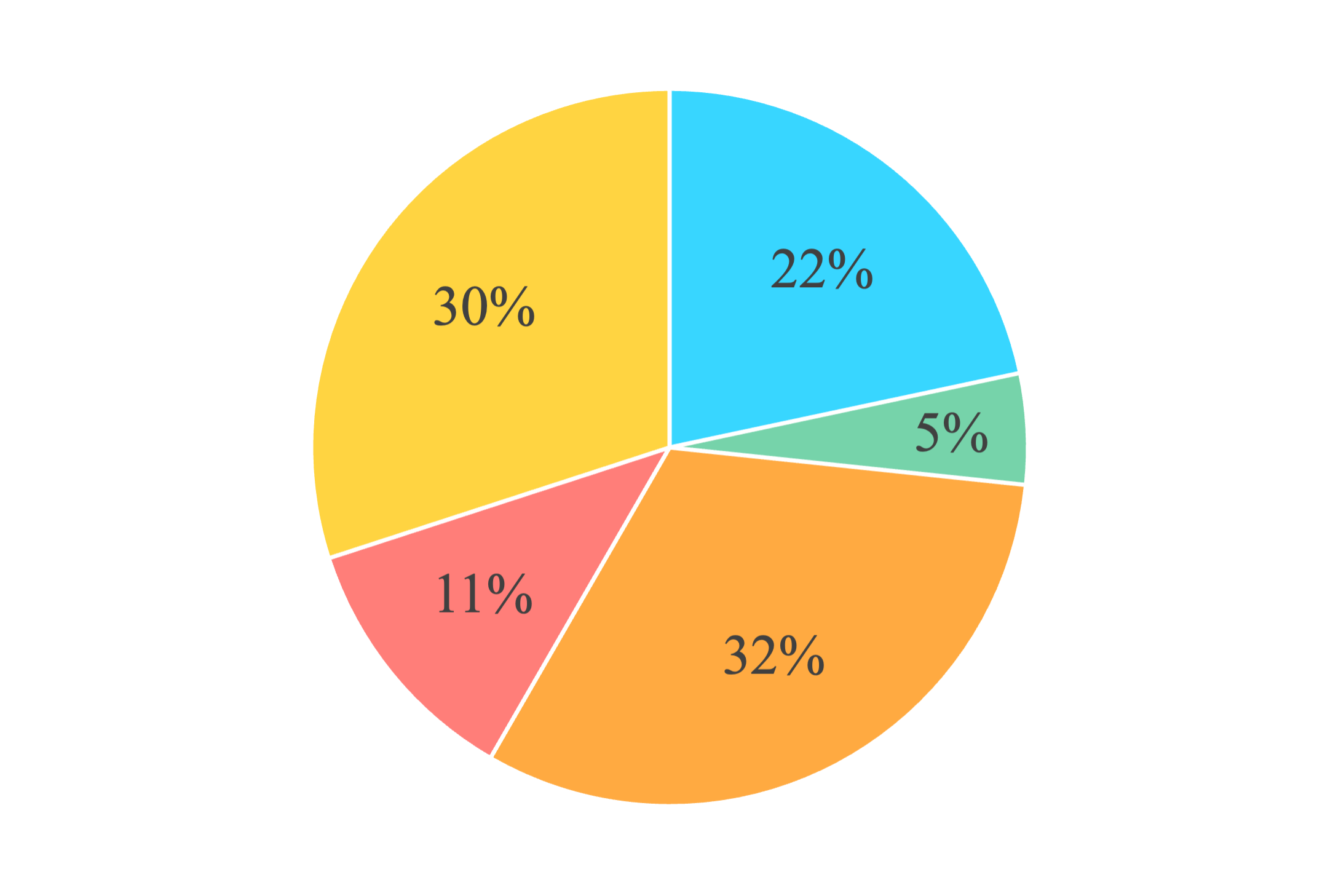}
        \caption{Selective Data}
        \label{fig:stage3_distribution}
    \end{subfigure}%
    \hfill
    \begin{subfigure}[b]{0.66\columnwidth}
        \centering
        \includegraphics[width=\textwidth]{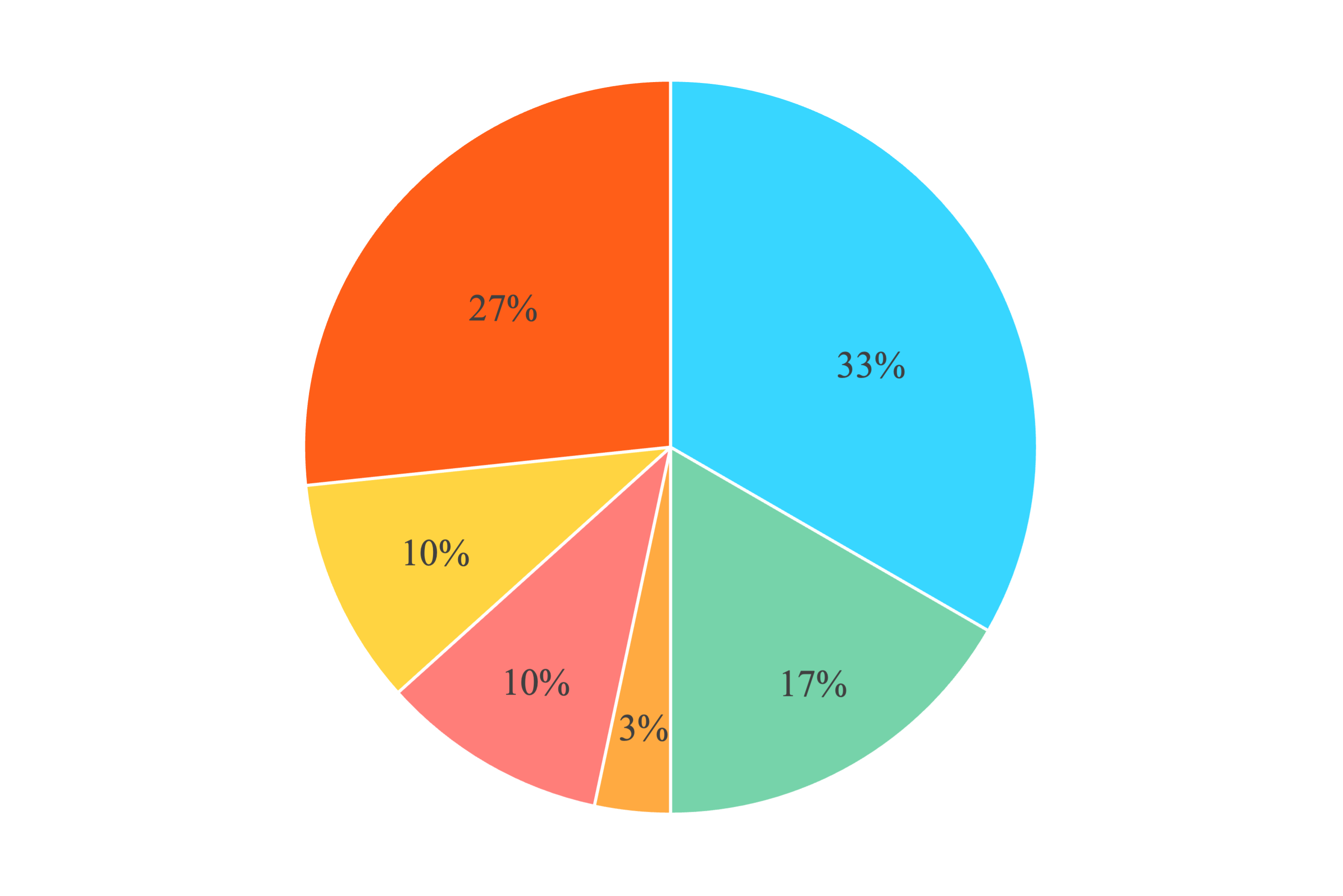}
        \caption{GPT4 \& Synthetic Data}
        \label{fig:stage4_distribution}
    \end{subfigure}%
    \hfill
    \begin{subfigure}[b]{1.5\columnwidth}
        \centering
        \includegraphics[width=\textwidth]{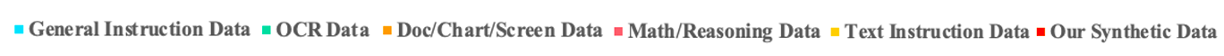}
        \label{fig:graphic_symbol}
    \end{subfigure}
    \caption{Distribution of Visual Instruction Data}
    \label{fig:distribution_of_visual_instruction_data}
\end{figure*}

For Image Caption Data, it consists entirely of General Instruction Data. But for Visual Instruction Data, due to its complex components, we further divide it into 3 parts, and visualize the data types of the three parts respectively, as shown in Figure~\ref{fig:distribution_of_visual_instruction_data}. The composition of Comprehensive Data and Selective Data is General Instruction Data, OCR Data, Doc/Chart/Screen Data, Math/Reasoning Data and Text Instruction Data, for GPT4 \& Synthetic Data, we add our Synthetic Data. Besides, we listed the sources, sizes, and types of all our data in Table~\ref{tab:data_composition}.


\begin{table*}[]
\begin{tabular}{|c|c|c|}
\hline
\textbf{Data Source}     & \textbf{Size}            & \textbf{Type}                                                                                                                                                                                            \\ \hline
Emu2 \citep{SunYCZZWGL0W24}                     & 10M                      & Caption                                                                                                                                                                                                  \\ \hline
LVIS-Instruct\citep{DBLP:conf/cvpr/GuptaDG19}            & 223K                     & General                                                                                                                                                                                                  \\ \hline
LLaVA-CC3M-Pretrain-595K\citep{li2024llava} & 595K                     & General                                                                                                                                                                                                  \\ \hline
Visdial\citep{DBLP:conf/cvpr/DasKGSYMPB17}                  & 116K                     & General                                                                                                                                                                                                  \\ \hline
Sharegpt4\citep{chen2023sharegpt4v}                & 3.2M                     & General                                                                                                                                                                                                  \\ \hline
STVQA\citep{DBLP:journals/ci/AgrawalJS24a}                    & 43K                      & General                                                                                                                                                                                                  \\ \hline
MMC-INST\citep{DBLP:conf/naacl/LiuWYCSCYY24}                 & 500K                     & Doc/Chart/Screen                                                                                                                                                                                         \\ \hline
MathV360K\citep{DBLP:journals/corr/abs-2406-17294}                & 338K                     & Math/Reasoning                                                                                                                                                                                           \\ \hline
MMC-Alignment\citep{DBLP:conf/naacl/LiuWYCSCYY24}            & 250K                     & Doc/Chart/Screen                                                                                                                                                                                         \\ \hline
DocReason\citep{DBLP:conf/cvpr/YeXYYHL0Z024}                & 26K                      & Doc/Chart/Screen                                                                                                                                                                                         \\ \hline
ALLaVA\citep{chen2024allava}                   & 1.7M                     & General                                                                                                                                                                                                  \\ \hline
Cocotext\citep{DBLP:journals/corr/HavardBR17}                 & 163K                     & General                                                                                                                                                                                                  \\ \hline
Docvqa\citep{DBLP:conf/wacv/MathewKJ21}                   & 16K                      & Doc/Chart/Screen                                                                                                                                                                                         \\ \hline
Geoqa+\citep{DBLP:conf/acl/ChenTQLLXL21}                   & 72K                      & Math/Reasoning                                                                                                                                                                                           \\ \hline
DocDownstream\citep{DBLP:conf/cvpr/YeXYYHL0Z024}            & 700K                     & Doc/Chart/Screen                                                                                                                                                                                         \\ \hline
Cambrian~\cite{DBLP:journals/corr/abs-2406-16860}                & 8.3M                     & \begin{tabular}[c]{@{}c@{}}General, General OCR, Math/Reasoning\\ Doc/Chart/Screen, Text Instruct\end{tabular}                                                                                        \\ \hline
DocStruct4M\citep{DBLP:conf/emnlp/WangZ0L20}              & 4M                       & General OCR, Doc/Chart/Screen                                                                                                                                                                            \\ \hline
LLaVA-onevision~\cite{DBLP:journals/corr/abs-2408-03326}          & 4M                       & \begin{tabular}[c]{@{}c@{}}General, General OCR, Math/Reasoning\\ Doc/Chart/Screen, Text Instruct\end{tabular}                                                                                                                                    \\ \hline
Docmatix\citep{DBLP:journals/corr/abs-2408-12637}                 & 1.2M                     & Doc VQA                                                                                                                                                                                                  \\ \hline
Infinity-Instruct~\cite{InfinityInstruct2024}       & 7M                       & Text Instruct                                                                                                                                                                                            \\ \hline
Our Synthetic Data       & 0.8M                     & \begin{tabular}[c]{@{}c@{}}Fine-grained Perception(single-instance)\\ Attribute Reasoning\\ Fine-grained Perception(Cross-instance)\\ Relation Reasoning\\ Coarse Perception, Logic Reasoning\end{tabular} \\ \hline
\end{tabular}
\centering
\caption{Data Source, Size and Type of Training Data}
\label{tab:data_composition}
\end{table*}

\section{Instruction Tagging System}
\label{appen:label}
\subsection{Coarse Perception}
\begin{itemize}
    \item Image Scene
    \begin{itemize}
        \item Identify structures
        \item Identify geographic location
        \item Identify weather condition
        \item Identify presence of people
        \item Identify event type
        \item Identify activity
        \item Identify location
        \item Identify time
        \item Identify buildings
        \item Identify people
        \item Other scene descriptions
        \item Identify background
        \item Identify diagram
        \item Identify action
        \item Identify season
        \item Identify vegetation type
        \item Other
        \item Identify objects in scene
        \item Identify overall theme
        \item Identify natural elements
        \item Identify objects
        \item Identify time of day
        \item Identify activities
        \item Identify number of people
        \item Identify environment type
        \item Count people
        \item Identify main subject
        \item Identify clothing
        \item Identify geometric properties
        \item Identify vegetation presence
        \item Identify animals
        \item Identify furniture
        \item Describe background
        \item Identify key elements
        \item Identify transportation
        \item Identify background details
        \item Identify presence of objects
        \item Identify natural environment scenery
        \item Other image scenes
        \item Identify stage
        \item Identify indoor scene
        \item Other image scene descriptions
        \item Identify temperature state
        \item Identify presence
        \item Describe scene
    \end{itemize}
    \item Image Quality
    \begin{itemize}
        \item Assess color and balance
        \item Assess focus
        \item Other image quality assessments
        \item Identify quality issues
        \item Assess brightness/ contrast
        \item Assess color
        \item Assess overall quality
        \item Assess lighting
        \item Assess overall clarity
        \item Assess composition
        \item Assess clarity
        \item Detect noise
        \item Assess sharpness
    \end{itemize}
    \item Image Topic
    \begin{itemize}
        \item Identify food
        \item Identify book-related content
        \item Identify animals
        \item Identify medical condition
        \item Identify geometric properties
        \item Identify people
        \item Identify portrait
        \item Identify objects
        \item Identify main subject
        \item Other image topics
        \item Identify text
        \item Identify event
        \item Identify diagram content
        \item Identify book
        \item Identify color
        \item Identify content
        \item Identify caption
        \item Identify infographic/ cartoon style
        \item Identify life cycle stage
        \item Identify chart content
        \item Identify image content
        \item Identify book content
        \item Identify vehicles
        \item Describe image
        \item Identify plant
        \item Identify sports
    \end{itemize}
    \item Image Emotion
    \begin{itemize}
        \item Detect overall emotion
        \item Other image emotion
        \item Read emotions from faces
    \end{itemize}
    \item Image Style
    \begin{itemize}
        \item Other image styles
        \item Identify image category
    \end{itemize}
\end{itemize}
\subsection{Fine-grained Perception (single-instance)}
\begin{itemize}
    \item Object Localization
    \begin{itemize}
        \item Locate object
        \item Determine coordinates
        \item Count objects
        \item Identify specific object
        \item Describe region
        \item Detect presence
        \item Provide bounding box
        \item Determine orientation
        \item Provide bounding box coordinates
        \item Count people
        \item Other object localization tasks
        \item Provide descriptions
        \item Count animals
        \item Provide region description
        \item Provide short description
        \item Identify region
        \item Other localization tasks
    \end{itemize}
    \item Attribute Recognition
    \begin{itemize}
        \item Recognize texture
        \item Recognize material
        \item Recognize pattern
        \item Recognize clothing
        \item Recognize geometric properties
        \item Recognize object presence
        \item Recognize appearance characteristics
        \item Recognize size
        \item Recognize objects
        \item Recognize color
        \item Other attributes
        \item Recognize formulas/tables/charts
        \item Recognize orientation
        \item Recognize shape
        \item Recognize category
        \item Count objects
    \end{itemize}
    \item OCR
    \begin{itemize}
        \item Recognize printed text
        \item Recognize text
        \item Transcribe text from image
        \item Extract text from image
        \item Recognize text in images
        \item Transcribe text in image
        \item Extract text from images
        \item Recognize formulas/ tables/ charts
        \item Key Information Extraction
        \item Transcribe text
        \item Other OCR tasks
    \end{itemize}
    \item Identify specific object
    \begin{itemize}
        \item direct
    \end{itemize}
    \item Detect presence
    \begin{itemize}
        \item direct
    \end{itemize}
\end{itemize}
\subsection{Fine-grained Perception (cross-instance)}
\begin{itemize}
    \item Spatial Relationship
    \begin{itemize}
        \item Determine relative position
        \item Determine spatial arrangement
        \item Other spatial relationships
        \item Determine coordinates
        \item Count objects
    \end{itemize}
    \item Action Recognition
    \begin{itemize}
        \item Recognize actions in video and text
        \item Recognize sequence of actions
        \item Recognize human-human interactions
        \item Recognize human actions
        \item Recognize animal actions
        \item Recognize human-object interactions
        \item Recognize actions
    \end{itemize}
    \item Attribute Comparison
    \begin{itemize}
        \item Compare text
        \item Other attribute comparison
        \item Compare preferences
        \item Compare ages
        \item Compare materials
        \item Compare values
        \item Compare material
        \item Compare shapes
        \item Compare shapes/ colors/ textures/ sizes
        \item Compare quantities
        \item Compare sizes
        \item Compare colors
    \end{itemize}
    \item Determine relative position
    \begin{itemize}
        \item direct
    \end{itemize}
\end{itemize}
\subsection{Relation Reasoning}
\begin{itemize}
    \item Social Relation
    \begin{itemize}
        \item Other social relations
        \item Identify family/ friendship/ professional/ hostile relationships
    \end{itemize}
    \item Physical Relation
    \begin{itemize}
        \item Identify spatial/ mechanical/ cause-effect relationships
        \item Identify cause-effect relationships
        \item Other physical relations
    \end{itemize}
    \item Nature Relation
    \begin{itemize}
        \item Other nature relations
    \end{itemize}
\end{itemize}
\subsection{Attribute Reasoning}
\begin{itemize}
    \item Identity Reasoning
    \begin{itemize}
        \item Other identity reasoning
        \item Predict occupation/ role/ social status
    \end{itemize}
    \item Function Reasoning
    \begin{itemize}
        \item Predict function of objects
        \item Other function reasoning
        \item New tag
    \end{itemize}
    \item Physical Property Reasoning
    \begin{itemize}
        \item Other physical properties
        \item Recognize geometric properties
        \item Other physical property reasoning
        \item Attribute Reasoning
        \item Recognize formulas/ tables/ charts
    \end{itemize}
\end{itemize}
\subsection{Logic Reasoning}
\begin{itemize}
    \item Structuralized Image-Text Understanding
    \begin{itemize}
        \item Parse tables
        \item Other image-text understanding
        \item Parse geometric diagrams
        \item Other Structuralized Image-Text Understanding
        \item Parse sales data
        \item Parse bar charts
        \item Parse line charts
        \item Parse text
        \item Other charts
        \item Parse other charts
        \item Parse diagrams
        \item Parse mathematical problem
        \item Parse bar/ pie/ line charts
        \item Parse formulas
        \item Parse function plots
        \item Parse charts
        \item Parse word problems
    \end{itemize}
    \item Future Prediction
    \begin{itemize}
        \item Predict trend/ social interaction/ physical movement/environmental changes
        \item Other future predictions
        \item Predict action sequence
        \item Action Prediction
        \item Predict series of actions
    \end{itemize}
\end{itemize}

\end{document}